\documentclass[lettersize,journal]{IEEEtran}
\usepackage{amsmath,amsfonts}
\usepackage{algorithmic}
\usepackage{algorithm}
\usepackage{array}
\usepackage[caption=false,font=normalsize,labelfont=sf,textfont=sf]{subfig}
\usepackage{textcomp}
\usepackage{stfloats}
\usepackage{url}
\usepackage{verbatim}
\usepackage{graphicx}
\usepackage{cite}
\usepackage{multirow}
\usepackage{booktabs}
\usepackage{makecell}
\usepackage{float}
\usepackage{bm}
\usepackage{bbm}
\usepackage[table]{ xcolor}
\usepackage{ulem}

\begin{document}

% \title{A Sample Article Using IEEEtran.cls\\ for IEEE Journals and Transactions}
\title{RobustMVS: Single Domain Generalized Deep Multi-view Stereo}

% \author{IEEE Publication Technology,~\IEEEmembership{Staff,~IEEE,}
%         % <-this % stops a space
% \thanks{This paper was produced by the IEEE Publication Technology Group. They are in Piscataway, NJ.}% <-this % stops a space
% \thanks{Manuscript received April 19, 2021; revised August 16, 2021.}}

\author{Hongbin Xu \ Weitao Chen \ Baigui Sun \ Xuansong Xie \ Wenxiong Kang\IEEEauthorrefmark{1}
\thanks{\IEEEauthorrefmark{1}Corresponding author.}
\thanks{Hongbin Xu is with the School of Automation Science and Engineering, South China University of Technology, Guangzhou, China, 510641. Email: hongbinxu1013@gmail.com}
\thanks{Weitao Chen, Baigui Sun, and Xuansong Xie are with the Damo Academy, Alibaba Group, Hanzhoug, China, 311121.}
\thanks{Wenxiong Kang is with the School of Auomation Science and Engineering, South China University of Technology, Guangzhou, China, 510641 and Pazhou Lab, Guangzhou, China, 510335. Email: auwxkang@scut.edu.cn.}
% \thanks{Manuscript received -; revised -.}
}

% The paper headers
\markboth{Journal of \LaTeX\ Class Files,~Vol.~14, No.~8, August~2021}%
{Shell \MakeLowercase{\textit{et al.}}: A Sample Article Using IEEEtran.cls for IEEE Journals}

% \IEEEpubid{0000--0000/00\$00.00~\copyright~2021 IEEE}
% Remember, if you use this you must call \IEEEpubidadjcol in the second
% column for its text to clear the IEEEpubid mark.
% \IEEEpubid{Copyright @ 20xx IEEE. Personal use of this material is permitted. However, permission to use this material for any other purposes must be obtained from the IEEE by sending an email to pubs-permissions@ieee.org.}
% \IEEEpubid{\begin{minipage}{\textwidth}\ \\[30pt] \centering
% 		Copyright \copyright 2024 IEEE. Personal use of this material is permitted. 
% 		However, permission to use this material for any other purposes must \\ be obtained 
% 		from the IEEE by sending an email to pubs-permissions@ieee.org.
% \end{minipage}}

\maketitle

\begin{abstract}
Despite the impressive performance of Multi-view Stereo (MVS) approaches given plenty of training samples, the performance degradation when generalizing to unseen domains has not been clearly explored yet. 
In this work, we {focus on} the domain generalization problem in MVS.
To evaluate the generalization results, we build a novel MVS domain generalization benchmark including synthetic and real-world datasets. In contrast to conventional domain generalization benchmarks, we consider a more realistic but challenging scenario, where only one source domain is available for training. 
The MVS problem can be analogized back to the feature matching task, and maintaining robust feature consistency among views is an important factor for improving generalization performance.
To address the domain generalization problem in MVS, we propose a novel MVS framework, namely RobustMVS\footnote{The code is released at: \url{https://github.com/ToughStoneX/Robust-MVS}.}. 
A Depth-Clustering-guided Whitening (DCW) loss is further introduced to preserve the feature consistency among different views, which decorrelates multi-view features from viewpoint-specific style information based on geometric priors from depth maps. 
The experimental results further show that our method achieves superior performance on the domain generalization benchmark\footnote{The MVS evaluation benchmark is released at: \url{https://github.com/ToughStoneX/MVS_Evaluation_Benchmark}}.
\end{abstract}


\section{Introduction}
\label{sec:introduction}
\IEEEPARstart{M}{ulti-view} Stereo (MVS) is a long-standing elementary problem in computer vision, which aims to reconstruct 3D scenarios from multi-view images.
In recent years, immense progress in this field has been witnessed since the advent of learning-based MVS.
Typically, a MVS network \cite{yao2018mvsnet} extracts multi-view image features with a shared CNN first and constructs cost volume via plane sweep algorithm \cite{collins1996space} where source images are reprojected to the reference view frustum. 
Afterward, this cost volume is further regularized by 3D CNN \cite{gu2020cascade, cheng2020deep, peng2022rethinking} or recurrent neural networks \cite{yao2019recurrent}.
Despite the amazing reconstruction performance and efficiency of learning-based MVS methods, the hunger for large-scale 3D annotations limits the application in realistic scenarios.
Consequently, some unsupervised MVS methods \cite{khot2019learning, huang2021m3vsnet, xu2021self} are further proposed to handle this issue without accessing ground truth.

\begin{figure}[t]
	\centering
	\includegraphics[width=\linewidth]{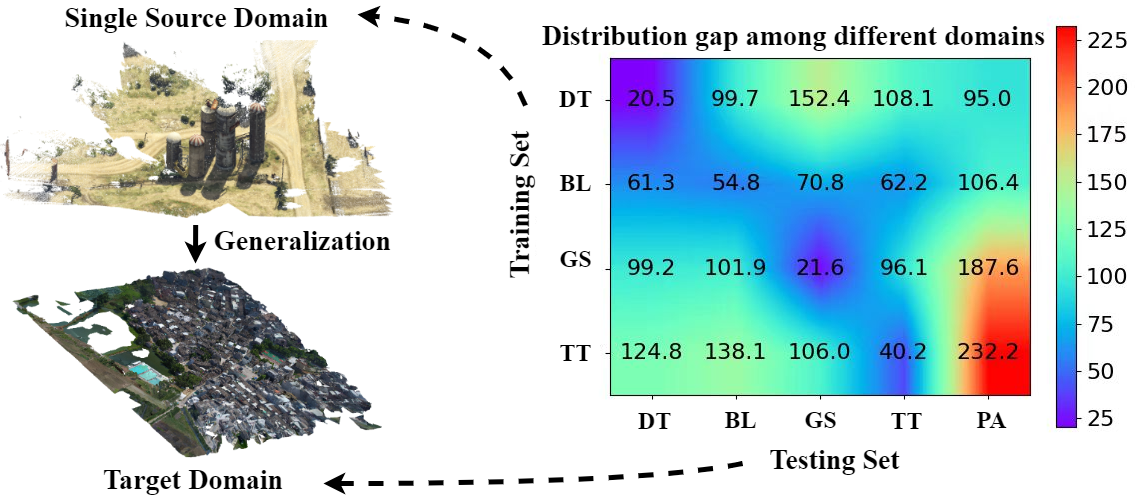}
	\caption{Illustration of the proposed MVS domain generalization task. A single source domain is selected for training, and further tested on target domains without finetuning. The MMD Distances among datasets are visualized on the right hand. MVS datasets: DTU (\textbf{DT}) \cite{jensen2014large}, BlendedMVS (\textbf{BL}) \cite{yao2020blendedmvs}, GTASFM (\textbf{GS}) \cite{wang2020flow}, PASMVS (\textbf{PA}) \cite{broekman2020pasmvs}, Tanks\&Temples (\textbf{TT}) \cite{knapitsch2017tanks}.}
	% \vspace{-0.4cm}
	\label{fig:domain_generalization_task}
\end{figure}

In previous state-of-the-art methods, much effort has been devoted to making more accurate and complete reconstruction results, but relatively little has been put into understanding the domain generalization problem in MVS.
% {
% In this paper, we first explore the problem of generalizing to unseen domains in MVS, especially the illuminance variations among different domains.
% Concretely, we consider a realistic and challenging setting, where only one source domain is available but generalizing to target domains without fine-tuning or adaptation, as shown in Fig. \ref{fig:domain_generalization_task}.
% }
{
In this paper, we aim to investigate the issue of generalizing to unseen domains in multi-view stereo (MVS), particularly the variations in illuminance among different domains. Specifically, we focus on a realistic and challenging scenario where there is only one source domain available but the goal is to generalize to target domains without any fine-tuning or adaptation, as illustrated in Fig. \ref{fig:domain_generalization_task}.
}
% In this paper, we first explore the problem of generalizing to unseen domains in MVS, considering a realistic and challenging setting, where only one source domain is available but generalizing to target domains without fine-tuning or adaptation, as shown in Fig. \ref{fig:domain_generalization_task}.
The major difficulties lie in the large domain differences (visualized via MMD distance in Fig. \ref{fig:domain_generalization_task}) among different MVS datasets, including factors like variations of color, illumination, and texture.
An interesting phenomenon can be found that the domain gap of \textbf{BL} $\rightarrow$ \textbf{TT} is much smaller than the one of \textbf{DT} $\rightarrow$ \textbf{TT}, which accounts for the better performance on \textbf{TT} test set when using \textbf{BL} as training set compared with \textbf{DT}, verifying the experiments in previous works\cite{ding2022transmvsnet, peng2022rethinking}.

\begin{figure*}[t]
	\centering
    \includegraphics[width=\linewidth]{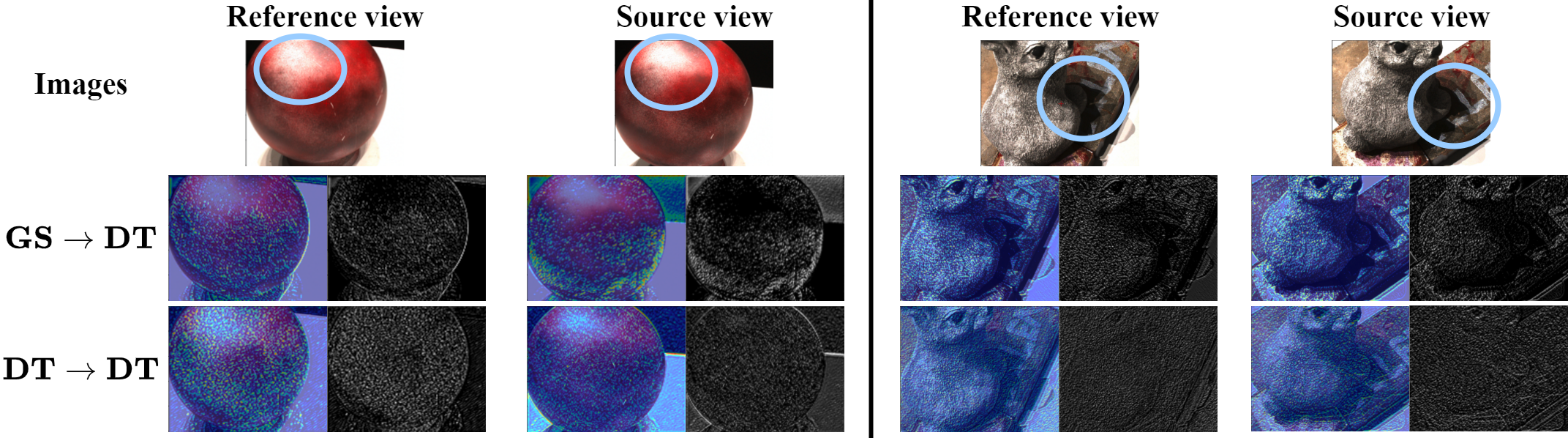}
%	\vspace{-0.6cm}
	\caption{Visualization of the activation map of CasMVSNet \cite{gu2020cascade} via Grad-Cam \cite{selvaraju2017grad}. \textbf{GS} $\rightarrow$ \textbf{DT} means training on domain \textbf{GS} and generalizing to unseen domain \textbf{DT}. \textbf{DT} $\rightarrow$ \textbf{DT} means training and testing on the same domain \textbf{DT}.}
	% \vspace{-0.4cm}
	\label{fig:visualization_grad_cam}
\end{figure*}

To handle the domain discrepancy, a common approach is to adapt the source domain distribution to the target domain distribution via domain adaptation  (DA) techniques \cite{ben2006analysis, ganin2015unsupervised, hoffman2018cycada,ye2023underwater}.
Whereas, it requires access to unlabeled samples in target domains, which limits their applicability in our MVS generalization task.
A better reference is the domain generalization (DG) technique \cite{muandet2013domain, li2018domain, li2018deep}, which aims to learn domain-invariant representations.
Furthermore, recent advances in the stereo matching task \cite{cai2020matching, zhang2020domain, shen2021cfnet, zhang2022revisiting, dai2022adaptive, lu2022a, pan2021multi} have been developed to handle the domain generalization issue via conducting feature-level alignment for obtaining domain-invariant features. 
The stereo matching task can be viewed as a simplified case of MVS, which suffers less from view-point variations and shares more common regions.
Instead of direct migration to the MVS case, we would like to take a step back first and revisit the bottleneck of generalization to unseen MVS domains.

To provide some inductive inspirations, we conduct a toy experiment that visualizes the feature activation map of an MVS network via modified Grad-CAM \cite{selvaraju2017grad}, as shown in Fig. \ref{fig:visualization_grad_cam}.
The first row shows the original multi-view images, and the following two rows respectively visualize the activation map when generalizing to unseen (\textbf{GS} $\rightarrow$ \textbf{DT}) and seen (\textbf{DT} $\rightarrow$ \textbf{DT}) domains.
The non-activated regions (the blue circled area in Fig. \ref{fig:visualization_grad_cam}) reveal the ignorance of MVS network during generalization.
Consequently, the bottleneck of domain generalization in MVS can be resorted to the robustness of cross-view matching features across domain-specific conditions of illumination, color, and reflectance.

In this paper, we propose a novel MVS framework, RobustMVS, with minimum modification to the backbone MVS network.
To handle the aforementioned issue that disturbs the feature consistency, we aim to suppress the domain-specific features and reserve the domain-invariant ones.
As a series of studies \cite{gatys2015texture, gatys2016image, pan2019switchable, luo2017learning, sun2016deep} claims, feature covariance implies domain-specific style such as color and illumination.
Based on this claim, the whitening transformation \cite{zhang2022revisiting, choi2021robustnet} is a technique designed for removing feature correlation and thus enhancing the invariance to style variation.
However, simply adopting the whitening transformation may not be applicable for MVS, because the whitening transformation decouples the feature style towards the whole image while MVS requires robust feature consistency among views in the local regions.
Hence, we introduce a novel Depth-Clustering-guided Whitening (DCW) loss in the RobustMVS framework.
With minor modifications to the MVS network, the last BN layer in each of the first 3 convolutional blocks is replaced with IN layer for normalization, as suggested by IBN-Net \cite{pan2018two}.
The whitening transformation is designed as a regularization loss without adding extra modules to the original network.
Our DCW loss firstly clusters the merged sparse point clouds generated from multi-view depth annotations to find out common regions which are spatially close.
Then the clusters are projected to each view which determines the corresponding local regions in different views.
By segmenting the local corresponding region (cluster) out, we warp the cross-view features to the same viewpoint via homography warping and calculate the feature covariance matrices on these correlated features.
Then the feature whitening constraints are applied to enhance the domain-invariant features in these local correlated regions.
Our experiments on MVS domain generalization task demonstrate the superior performance of the proposed framework.

The main contributions include the following:

% \begin{itemize}[itemsep=2pt,topsep=0pt,parsep=0pt]
%	\setlength{\itemsep}{0pt}
%	\setlength{\parsep}{0pt}
%	\setlength{\parskip}{0pt}
%	\setlength{\topsep}{0pt}
%	\setlength{\partopsep}{0pt}
\begin{itemize}
	% \item {We introduce the domain generalization problem in MVS}, and build customized domain generalization MVS benchmark for evaluation.
    % \item {We systematically investigate the domain generalization problem in MVS, and extend the domain generalized MVS benchmarks to 5 datasets for evaluation.}
    \item {We systematically investigate the domain generalization problem in MVS and propose a new domain generalized benchmark across multiple datasets in MVS. Different from the original ones, the evaluation protocols (\textbf{BL}/\textbf{GS}/\textbf{GS}) are specially customized for measuring the performance of 3D reconstruction.}
    \item We propose a novel domain generalized MVS framework, RobustMVS. By applying minor modifications to the MVS network and training with the proposed Depth-Clustering-guided Whitening (DCW) loss, the RobustMVS framework can improve the generalization ability with negligible computational cost.
	\item The domain generalized evaluation results show the superiority of our approach over existing approaches in both a qualitative and quantitative manner.
\end{itemize}

\section{Related Work}
\label{sec:related}

In this section, we briefly introduce the learning-based Multi-view Stereo (MVS) methods that are highly related to our paper.
In addition, we also review some related work of Domain Generalization methods to explore the single domain generalization classification task and their relationship with our work.

\subsection{Learning-based MVS}

With the blessing of large-scale MVS datasets like DTU \cite{jensen2014large} and BlendedMVS \cite{yao2020blendedmvs}, learning-based MVS methods are powered by deep convolutional networks \cite{yao2018mvsnet}, recurrent networks \cite{yao2019recurrent}, and Transformers \cite{ding2022transmvsnet}, achieving much better reconstruction accuracy and completeness than {traditional MVS methods \cite{galliani2015massively,campbell2008using,furukawa2010accurate}.}
The pioneering work of MVSNet \cite{yao2018mvsnet} projects the feature maps of multi-view images extracted by CNN to the same reference view frustum, and builds a 3D cost volume to model the matching cost on each hypothetical depth plane.
3D CNN is then used for cost aggregation, and the output per-view depth maps are fused to construct a 3D point cloud.
Due to the high memory and computation cost of MVSNet, lots of efforts have been proposed which can be divided into recurrent MVS networks and coarse-to-fine MVS networks.
The recurrent methods \cite{yan2020dense, yao2019recurrent, wei2021aa} use the Recurrent Neural Network (RNN)  to regularize the cost volume as an alternative to CNNs.
In this way, the computational efficiency is sacrificed in exchange for reducing memory cost.
% The recurrent methods utilize RNNs to propagate the features of 3D cost volume among different depth hypotheses, and the coarse-to-fine methods separate the single large cost volume into multi-stage cascaded cost volumes which narrow the depth range of previous prediction.
When the coarse-to-fine methods \cite{yang2020cost, cheng2020deep, su2022uncertainty, gu2020cascade, zhao2023epxloring} are adopted, the cost volume at the coarsest stage is constructed by uniformly sampling depth planes \cite{zhang2021farther} of all depth hypotheses.
Then the depth map is refined iteratively by building a more compact cost volume according to the estimated depth in previous stage as an initialization. 
CVP-MVSNet \cite{yang2020cost} proposes to adaptively determine the local depth search ranges by calculating the mean depth interval corresponding to 0.5 pixel distance on the epipolar line of the closest source images.
UCSNet \cite{cheng2020deep} utilizes the variance of the depth probability volume and depth residual to determine the pixel-wise depth interval when sampling depth hypotheses.
UGNet \cite{su2022uncertainty} enables the MVS network to perceive the uncertainty, that is further used to guide the depth hypothesis sampling and build more compact cost volumes without redundancy.
CasMVSNet \cite{gu2020cascade} relies on the hyperparameters obtained from practice to determine the depth interval in coarse-to-fine cost volumes.
Furthermore, some recent works \cite{wang2022mvster, ding2022transmvsnet} attempt to embed MVS network with Transformer \cite{vaswani2017attention} backbones, achieving impressive improvement compared with previous methods.
Another recent strand of work is unsupervised MVS \cite{khot2019learning, huang2021m3vsnet, xu2021self, xu2021digging,mallick2020learning}, which designs self-supervised loss based on photometric consistency \cite{chen2021fixing,wei2021iterative} to train the MVS networks, producing competitive performance compared to supervised methods.
Despite the promising results of learning-based MVS methods, the generalization problem in MVS has not been systematically explored yet.
% Though the generalization experiments to Tanks\&Temples dataset is widely used in previous methods, the source data comes from DTU dataset, BlendedMVS dataset or both datasets.
{Though the generalization experiments to Tanks\&Temples are widely used in previous methods, the experimental settings might vary because some methods \cite{yao2018mvsnet, yao2019recurrent} adopt DTU as the source dataset while others utilize BlendedMVS \cite{zhang2020visibility} or multiple datasets \cite{ding2022transmvsnet}.
As Fig. \ref{fig:domain_generalization_task} shows, the distribution gap of BlendedMVS is much smaller than DTU, which might affect the final generalization performance when different source data settings are adopted.}
The exact influence of distribution among different MVS datasets has not been systematically investigated yet.
More challenging but worthwhile, we introduce a novel domain generalization problem in MVS and aim to handle it in this paper.

\subsection{Domain Generalization in Stereo Vision}

Domain generalization \cite{muandet2013domain, li2018domain, li2018deep} aims to learn from multiple source domains but generalize to unseen target domains.
These techniques have also been extended to the era of stereo matching \cite{cai2020matching, zhang2020domain, shen2021cfnet, zhang2022revisiting}, which is a special case of MVS with horizontally aligned camera poses.
DSMNet \cite{zhang2020domain} adopts a domain normalization layer to reduce shifts of domain-specific styles in images, and CFNet \cite{shen2021cfnet} proposes a fused cost volume representation that  is robust to domain variations.
FCStereo \cite{zhang2022revisiting} involves selective whitening constraints to stereo matching to suppress the domain-sensitive features.
The domain generalized stereo matching methods provide a reliable reference, and we extend the state-of-the-art methods to MVS network for comparison in Section \ref{sec:exp}.

\section{Preliminary}
\label{sec:preliminary}

In this section, we briefly introduce the concept of feature whitening transformation operation and its application to domain generalized tasks like Semantic segmentation and Stereo Matching in Section \ref{sec:preliminary:whitening_loss}. 
Then we discuss the limitation of the Whitening loss towards our MVS task in Section \ref{sec:preliminary:limitation}.

\subsection{Whitening Transformation and Whitening Loss}
\label{sec:preliminary:whitening_loss}
% \noindent\textbf{Whitening Transformation and Whitening Loss:}

Denote that $\mathbf{F} \in \mathbb{R}^{C \times {N_{HW}}}$ is a flattened feature map, whose number of channels is $C$ {and $N_{HW}=H*W$}.
$H$ and $W$ respectively represent the height and width.
As previous studies \cite{gatys2015texture, gatys2016image, pan2019switchable, luo2017learning, sun2016deep} claim, feature covariance matrix $\mathbf{F} \cdot \mathbf{F}^T \in \mathbb{R}^{C \times C}$ contains the visual style information of the whole image.
Based on this intuition, whitening transformation (WT) is a linear transformation that aims to make the variance term of each channel in the feature map equal to one and the covariance term among different channels equal to zero.
More concretely, it aims to find a whitened feature map $\tilde{\mathbf{F}} \in \mathbb{R}^{C \times {N_{HW}}}$ that satisfies $\tilde{\mathbf{F}} \cdot \tilde{\mathbf{F}}^T = {N_{HW}} \cdot \mathbf{I} \in \mathbb{R}^{C \times C}$, where $\mathbf{I}$ is an identity matrix.

The mean vector of the feature map can be calculated as:
\begin{equation}
%	\setlength{\abovedisplayskip}{2pt}
%	\setlength{\belowdisplayskip}{2pt}
%	\abovedisplayshortskip=0pt
%	\belowdisplayshortskip=0pt
%	\abovedisplayskip=0pt
%	\belowdisplayskip=0pt
	% \small
	\bm{\mu} = \frac{1}{{N_{HW}}} \mathbf{F} \cdot \mathbf{1}, \mathbf{1} \in \mathbb{R}^{{N_{HW}} \times 1}
	\label{eq1}
\end{equation}

Then the covariance matrix can be computed as:
\begin{equation}
	% \small
	\mathbf{\Sigma} = \frac{1}{{N_{HW}}} (\mathbf{F} - \bm{\mu} \cdot \mathbf{1}^T) (\mathbf{F} - \bm{\mu} \cdot \mathbf{1}^T)^T, \mathbf{1} \in \mathbb{R}^{1 \times {N_{HW}}}
	\label{eq2}
\end{equation}

Afterwards, the solution of whitened feature map is:
\begin{equation}
	% \small
	\tilde{\mathbf{F}} = \mathbf{\Sigma}^{-\frac{1}{2}} (\mathbf{F} - \bm{\mu} \cdot \mathbf{1}^T), \mathbf{1} \in \mathbb{R}^{1 \times {N_{HW}}}
	\label{eq3}
\end{equation}

The covariance matrix $\mathbf{\Sigma}$ can be eigen-decomposed to $\mathbf{Q} \mathbf{\Lambda} \mathbf{Q}^T$, where $\mathbf{Q} \in \mathbb{R}^{C \times C}$ is the orthogonal matrix of eigenvectors, and $\mathbf{\Lambda} \in \mathbb{R}^{C \times C}$ is the diagonal matrix that contains eigenvalues.
The inverse square root of the covariance matrix can be calculated with $\mathbf{\Sigma}^{-\frac{1}{2}} = \mathbf{Q} \mathbf{\Lambda}^{-\frac{1}{2}} \mathbf{Q}^T$, substituting the corresponding term in Eq. \ref{eq3}.

Since computing whitening transformation matrix analytically during inference is computationally expensive, GDWCT \cite{cho2019image} converts the whitening constraints to a regularization loss on feature maps:
\begin{equation}
	% \small
	L_{\text{whiten}} = \mathbb{E}[ \| \mathbf{\Sigma} - \mathbf{I} \|_1 ]
	\label{eq4}
\end{equation}

The whitening loss $L_{\text{whiten}}$ and the costful computation process of calculating eigendecomposition are only used during training, while no extra computation is conducted during inference.

% \noindent\textbf{Limitations of Whitening Loss in MVS:}
\subsection{Limitations of Whitening Loss in MVS}
\label{sec:preliminary:limitation}

With the nature of removing style information \cite{li2017universal}, the whitening constraints can be appended to the domain generalization task for learning domain-invariant representations.
Whereas the domain-specific and domain-invariant features are simultaneously encoded in the covariance matrix of feature maps, they may be diminished by whitening loss at the same time.
Some extensions \cite{choi2021robustnet, zhang2022revisiting} handle this by selecting the domain-specific and domain-invariant terms of the covariance matrix via its sensitivity to changes like photometric transformations.
% However, these whitening losses are operated on the whole image to regularize the style difference among all pixels in the whole image, which may not fit our MVS setting, whose core is the feature-matching correspondence.
{
However, these whitening losses are operated on the whole image to regularize the style difference among all pixels in the whole image.
It is in conflict with the core target of MVS, which concentrates on the correspondence among pixels in local regions rather than the whole image.}
As the discussion in Section \ref{sec:introduction} of Fig. \ref{fig:visualization_grad_cam} reveals, the robustness of local corresponding features among views towards changes is the bottleneck when generalizing to unseen domains.
Hence, instead of concentrating on the whole image with whitening loss, we propose a novel Depth-Clustering-guided Whitening (DCW) loss that applies whitening loss to the local corresponding regions of all views.

\section{Method}
\label{sec:method}

In this section, we design a novel RobustMVS framework to handle the domain generalized problem in MVS.
In Section \ref{sec:method:problem}, we investigate the potential challenges of the domain generalized problem in MVS.
Then, we introduce the adopted backbone network of the MVS framework in Section \ref{sec:method:backbone}.
In Section \ref{sec:method:conv_block_with_bn}, we introduce the modification of MVS networks to support the feature whitening operations.
Afterwards, we introduce the Depth-Clustering-guided Whitening (DCW) loss of this work which aims to conduct feature whitening selectively on the local corresponding regions among different views. 

% Given a pair of multi-view images with $N$ views, the image on reference view is noted as $I_{\text{ref}} = I_1$, and the the source view image is denoted as $I_{\text{src}} = \{ I_v \}_{n=2}^{N}$.
% In the same way, the intrinsic and extrinsic matrix is written as: $K_{\text{ref}}$/$K_{\text{src}}$, $T_{\text{ref}}$/$T_{\text{src}}$.
% Accordingly, the ground truth depth map is written as: $D_{\text{ref}}$/$D_{\text{src}}$.

\begin{figure*}[t]
	\centering
    \includegraphics[width=\linewidth]{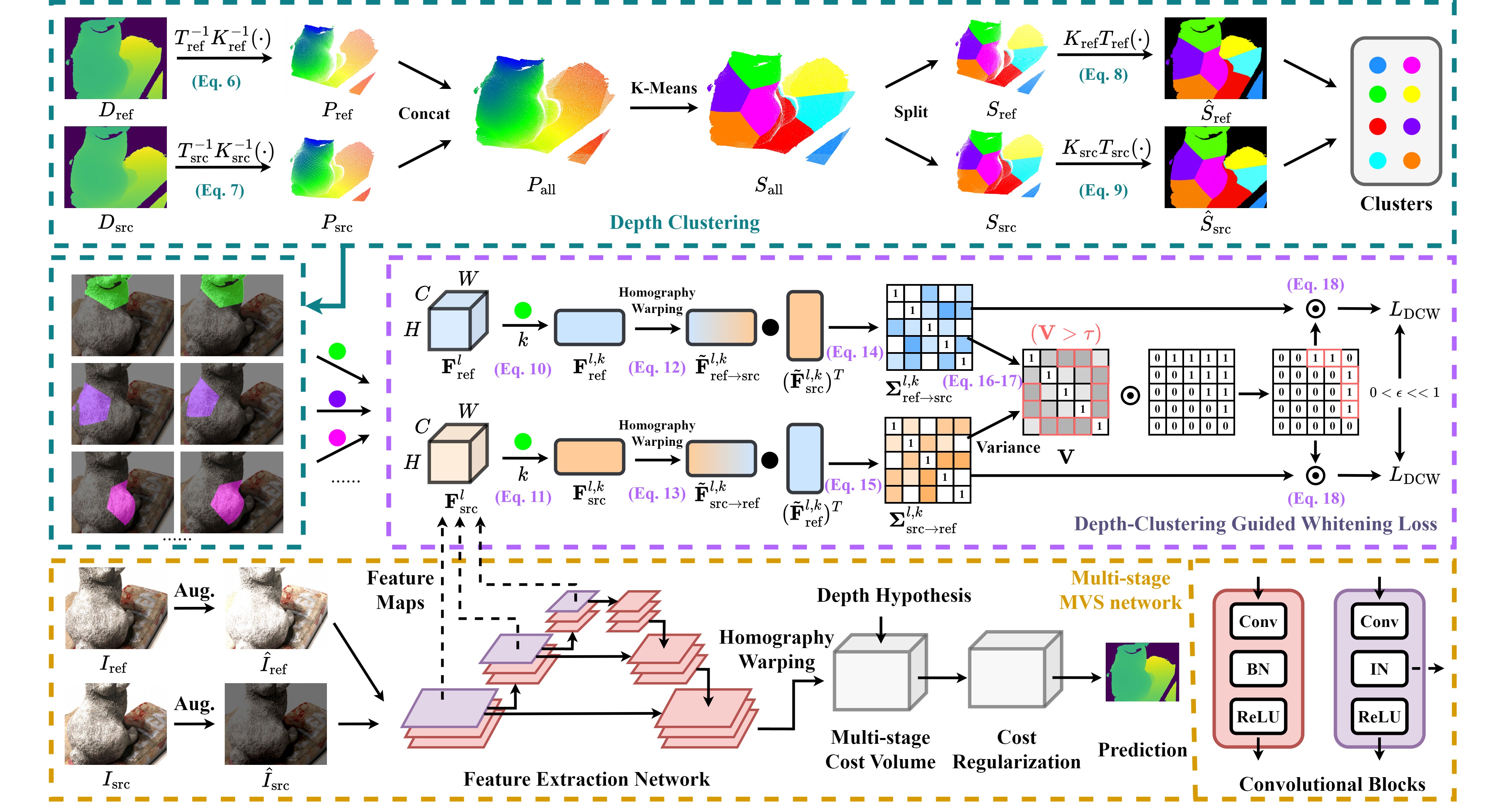}
	%	\vspace{-0.6cm}
	\caption{Illustration of our proposed RobustMVS framework. Aug. means random data augmentation. {
 % The blue dashed box shows the clustering process with the depth prior. It can extract the local corresponding regions by clustering, and the clusters are further fed to the proposed Depth-Clustering Guided Whitening loss shown in the purple dashed box. The orange dashed box at the bottom shows the basic architecture of the adopted MVS network whose first 3 BN convolution blocks are replaced to the IN-based convolution blocks.
 The blue dashed box represents the clustering process with the depth prior. This process can extract local corresponding regions by clustering, and the clusters are further fed to the proposed Depth-Clustering Guided Whitening loss shown in the purple dashed box. The orange dashed box at the bottom represents the basic architecture of the adopted MVS network, in which the first 3 BN convolution blocks are replaced with the IN-based convolution blocks.
 }}
	% \vspace{-0.4cm}
	\label{fig:framework}
\end{figure*}

\subsection{Problem Statement and Notation}
\label{sec:method:problem}

We first introduce the notations and definition of the domain generalization problem in MVS.
% As shown in Fig. \ref{fig:domain_generalization_task}, we select a single source domain from the training sets of \textbf{DT}, \textbf{BL}, \textbf{GS}, \textbf{TT}, and the target domain from the testing sets of \textbf{DT}, \textbf{BL}, \textbf{GS}, \textbf{TT}, and \textbf{PA}.
The goal of our MVS domain generalization task is to learn a model that is trained only using data from a source domain that can generalize well to the unseen target domain.
In the source domain, given a pair of multi-view images with $N$ views, the image on reference view is noted as $I_{\text{ref}} = I_1$, and the source view image is denoted as $I_{\text{src}} = \{ I_v \}_{n=2}^{N}$.
In the same way, the intrinsic and extrinsic matrix are written as: $K_{\text{ref}}$/$K_{\text{src}}$, $T_{\text{ref}}$/$T_{\text{src}}$.
Accordingly, the ground truth depth map is written as $D_{\text{ref}}$/$D_{\text{src}}$.

\subsection{Backbone}
\label{sec:method:backbone}

As shown in the third row of Fig. \ref{fig:framework}, the backbone MVS network of RobustMVS is a multi-stage MVS network that follows a coarse-to-fine pipeline as discussed in Section \ref{sec:related}.
In default, we choose the foundational CasMVSNet \cite{gu2020cascade} as the backbone.
On each view, the output feature map of the feature extraction network is projected to the reference view via homography warping.
Based on the given depth hypothesis either from initial settings or from the prediction of the previous stage, the cost volume is built to measure the similarity of corresponding points among views.
Then the cost volume is fed to a 3D CNN in the cost regularization process, and finally outputs the predicted depth.

\subsection{Convolutional Block with Normalization}
\label{sec:method:conv_block_with_bn}

The optimization of the aforementioned whitening loss in Eq. \ref{eq4} can be separated into two parts: optimization on the diagonal and off-diagonal of the covariance matrix. 
Denote that $F_i$ is channel $i$ of feature map $F$.
For the diagonal constraints, its loss can be written as: $\| \mathbf{\Sigma} - 1 \|_1 = \| \frac{F_i F_i^T}{{N_{HW}}} - 1 \|_1$.
For the off-diagonal constraints, its loss can be written as: $\| \mathbf{\Sigma} \|_1 = \| \frac{F_i F_j^T}{{N_{HW}}}\|_1$.
The former constraint forces the scale of channel $F_i$ to be $\frac{1}{{N_{HW}}}$, while the latter constraint forces the target to be zero.
This conflict may confuse the training process.
Hence, a simple solution to handle this issue is to standardize the feature map with Instance Normalization \cite{ulyanov2016instance} to ensure a fixed scale for each channel $F_i$:
\begin{equation}
	% \small
	\mathbf{F}_s = (\text{diag}(\mathbf{\Sigma}))^{-\frac{1}{2}} \odot (\mathbf{F} - \bm{\mu} \cdot \mathbf{1}^T), \mathbf{1} \in \mathbb{R}^{1 \times {N_{HW}}}
	\label{eq5}
\end{equation}
where $\odot$ is an element-wise multiplication.
Then the normalized feature map can be used to compute the covariance matrix  Eq. \ref{eq2} and the whitening loss via Eq. \ref{eq4}

The summarized principle is that the whitening loss should be computed on the standardized output of an Instance Normalization layer.
With this in mind, we replace the Batch Normalization layer with the Instance Normalization layer when the output feature map is used to calculate the whitening loss, as shown in Fig. \ref{fig:framework}.

\subsection{Depth-Clustering-guided Whitening Loss}

%Following the discussion about Fig. \ref{fig:visualization_grad_cam} in Section \ref{sec:intro}, we aim to strengthen the robustness of cross-view feature consistency.
Since the core of MVS is the feature-matching task, which concentrates on the correspondence in local regions.
It is inappropriate to directly append whitening constraints to features in the whole image, because it may diminish specific style patterns which is beneficial for feature consistency in local regions.
Hence, we propose a novel Depth-Clustering-guided Whitening (DCW) loss to conduct feature whitening selectively on the local corresponding regions among different views.
In this section, we first introduce the procedure of segmenting the local correspondence via clustering with depth priors in Sec. \ref{sec:clustering_with_depth}, and then expound the feature whitening on cross-view corresponding features by homography warping in Sec. \ref{sec:homography_whitening}.

\subsubsection{Clustering with Depth Priors}
\label{sec:clustering_with_depth}

As shown in the first row of Fig. \ref{fig:framework}, the ground truth depth maps of reference view ($D_{\text{ref}} \in \mathbb{R}^{H \times W}$) and source view ($D_{\text{src}} \in \mathbb{R}^{H \times W}$) are first projected back into 3D space.
\begin{equation}
	% \small
	P_{\text{ref}}^i = T_{\text{ref}}^{-1} K_{\text{ref}}^{-1} D_{\text{ref}}(p_{\text{ref}}^i) p_{\text{ref}}^i
	\label{eq6}
\end{equation}
\begin{equation}
	% \small
	P_{\text{src}}^j = T_{\text{src}}^{-1} K_{\text{src}}^{-1} D_{\text{src}}(p_{\text{src}}^j) p_{\text{src}}^j
	\label{eq7}
\end{equation}
where $i$ and $j$ are respectively the index of valid points in $D_{\text{ref}}$ and $D_{\text{src}}$, and $p_{\text{ref}}$ and $p_{\text{src}}$ represent the 2D image coordinates.
Denote that the number of valid pixels in the reference view is $M_{\text{ref}}$ and the one in the source view is $M_{\text{src}}$.
The projected point cloud on reference view is $P_{\text{ref}} \in \mathbb{R}^{M_{\text{ref}} \times 3}$, and the point cloud on source view is $P_{\text{src}} \in \mathbb{R}^{M_{\text{src}} \times 3}$.

Then the point clouds $P_{\text{ref}}$ and $P_{\text{src}}$ are concatenated together to get a fused point cloud $P_{\text{all}} \in \mathbb{R}^{(M_{\text{ref}} + M_{\text{src}}) \times 3}$.
Afterward, K-Means \cite{hartigan1979algorithm} is adopted to cluster the spatially neighboring regions of the fused point cloud $P_{\text{all}}$.
The number of clusters is defined as {$K_{\text{clu}}$}, and the output clustered point cloud is $S_{\text{all}} \in \mathbb{R}^{(M_{\text{ref}} + M_{\text{src}}) \times {K_{\text{clu}}}}$.
The {$K_{\text{clu}}$} channels of point cloud $S_{\text{all}}$ use one-hot encoding to represent the {$K_{\text{clu}}$} clusters.
Each of the {$K_{\text{clu}}$} clusters in point cloud $S_{\text{all}}$ represents the spatially neighboring regions that bind the matching consistency of local regions between different views.
Afterward, point cloud $S_{\text{all}}$ can be split back into clusters on reference view and source view: $S_{\text{ref}} \in \mathbb{R}^{M_{\text{ref}} \times {K_{\text{clu}}}}$ and $S_{\text{src}} \in \mathbb{R}^{M_{\text{src}} \times {K_{\text{clu}}}}$.
Note that the {$K_{\text{clu}}$} clusters in $S_{\text{ref}}$ and $S_{\text{src}}$ are corresponding regions that inherently model the matching consistency among views.

The clustered point clouds can then be projected into the 2D image space:
\begin{equation}
	% \small
	\tilde{S}_{\text{ref}} (p_{\text{ref}}^i) = S_{\text{ref}} (P_{\text{ref}}^i), p_{\text{ref}}^i = \pi (K_{\text{ref}} T_{\text{ref}} P_{\text{ref}}^i)
\end{equation}
\begin{equation}
	% \small
	\tilde{S}_{\text{src}} (p_{\text{src}}^j) = S_{\text{src}} (P_{\text{src}}^j), p_{\text{src}}^j = \pi (K_{\text{src}} T_{\text{src}} P_{\text{src}}^j)
\end{equation}
where $\pi ([x,y,z]^T) = [x/z,y/z,1]$, $i$ and $j$ are respectively the index of valid points which can be projected into 2D image.
$\tilde{S}_{\text{ref}} \in \mathbb{R}^{H \times W \times {K_{\text{clu}}}}$ and $\tilde{S}_{\text{src}} \in \mathbb{R}^{H \times W \times {K_{\text{clu}}}}$ are 2D clustered segmentation maps that share common clusters among different views.

\subsubsection{Homography Whitening}
\label{sec:homography_whitening}

Denote that $\mathbf{F}_{\text{ref}}^l \in \mathbb{R}^{{N_{HW}} \times C}$ and $\mathbf{F}_{\text{src}}^l \in \mathbb{R}^{{N_{HW}} \times C}$ are the output feature maps on the $l$-th layer, which is normalized with Instance Normalization following Eq. \ref{eq5}.
As shown in Fig. \ref{fig:framework}, each feature map $\mathbf{F}_s^l$ is filtered with each of the {$K_{\text{clu}}$} clusters to find locally consistent regions:
\begin{equation}
	% \small
	\tilde{\mathbf{F}}_{\text{ref}}^{l, k} = \mathbf{F}_{\text{ref}}^l \odot \mathbb{1}(\tilde{S}_{\text{ref}}^k = 1)
\end{equation}
\begin{equation}
	\tilde{\mathbf{F}}_{\text{src}}^{l, k} = \mathbf{F}_{\text{src}}^l \odot \mathbb{1}(\tilde{S}_{\text{src}}^k = 1)
\end{equation}
where $\odot$ is an element-wise multiplication.
$\tilde{\mathbf{F}}_{\text{ref}}^{l, k} \in \mathbb{R}^{M_{ref}^k \times C}$ and $\tilde{\mathbf{F}}_{\text{src}}^{l, k} \in \mathbb{R}^{M_{src}^k \times C}$ are the pixel-wise features of cluster $k$ on the reference view and the source view.
$M_{ref}^k$ and $M_{src}^k$ are the numbers of valid pixels in cluster $K$ on reference view and source view.

Since $M_{ref}^k$ and $M_{src}^k$ are probably unequal, they can not be directly used to calculate the covariance matrix that has to be a square matrix.
Hence, we reproject $\tilde{\mathbf{F}}_{\text{ref}}^{l, k}$ to the source view and $\tilde{\mathbf{F}}_{\text{src}}^{l, k}$ to the reference view via homography warping:
\begin{equation}
	% \small
	\tilde{\mathbf{F}}_{\text{ref} \rightarrow \text{src}}^{l, k}[p_{\text{src}}] = \tilde{\mathbf{F}}_{\text{ref}}^{l, k} [\pi(K_{\text{ref}} T_{\text{ref}} T_{\text{src}}^{-1} K_{\text{src}}^{-1} p_{\text{src}} )]
\end{equation}
\begin{equation}
	% \small
	\tilde{\mathbf{F}}_{\text{src} \rightarrow \text{ref}}^{l, k}[p_{\text{ref}}] = \tilde{\mathbf{F}}_{\text{src}}^{l, k} [\pi(K_{\text{src}} T_{\text{src}} T_{\text{ref}}^{-1} K_{\text{ref}}^{-1} p_{\text{ref}} )]
\end{equation}
where $\tilde{\mathbf{F}}_{\text{src} \rightarrow \text{ref}}^{l, k} \in \mathbb{R}^{M_{ref}^k \times C}$ and $\tilde{\mathbf{F}}_{\text{ref} \rightarrow \text{src}}^{l, k} \in \mathbb{R}^{M_{src}^k \times C}$ are the warped features via bilinear sampling to ensure the backward propagation of gradient during training.

Then we can calculate the covariance matrix: 
\begin{equation}
	% \small
	\mathbf{\Sigma}_{\text{ref} \rightarrow \text{src}}^{l,k} = (\tilde{\mathbf{F}}_{\text{src}}^{l, k})^T \tilde{\mathbf{F}}_{\text{ref} \rightarrow \text{src}}^{l, k}
\end{equation}
\begin{equation}
	% \small
	\mathbf{\Sigma}_{\text{src} \rightarrow \text{ref}}^{l,k} = (\tilde{\mathbf{F}}_{\text{ref}}^{l, k})^T \tilde{\mathbf{F}}_{\text{src} \rightarrow \text{ref}}^{l, k}
\end{equation}
where the covariance matrix $\mathbf{\Sigma}_{\text{ref} \rightarrow \text{src}}^{l,k}$ and $\mathbf{\Sigma}_{\text{src} \rightarrow \text{ref}}^{l,k}$ have the resolution of $C \times C$.

Note that the original input of the reference image and source image are randomly augmented with photometric transformations as shown in the left-bottom of Fig. \ref{fig:framework}.
Here, the covariance matrix $\mathbf{\Sigma}_{\text{ref}  \rightarrow \text{src}}^{l,k}$ and $\mathbf{\Sigma}_{\text{src}  \rightarrow \text{ref}}^{l,k}$ models the style information on the local region of the same cluster $k$.
With the random data augmentation, we would like to find out which elements of the covariance matrix is sensitive to changes, and suppress them with whitening loss.
Formally, the variance matrix between these covariance matrices is calculated as follows:
\begin{equation}
	% \small
	\bm{\mu}_{\mathbf{\Sigma}} = \frac{\mathbf{\Sigma}_{\text{ref} \rightarrow \text{src}}^{l,k} + \mathbf{\Sigma}_{\text{src}  \rightarrow \text{ref}}^{l,k}}{2}
\end{equation}
\begin{equation}
	% \small
	\mathbf{V} = \frac{(\mathbf{\Sigma}_{\text{ref}  \rightarrow \text{src}}^{l,k} - \bm{\mu}_{\mathbf{\Sigma}})^2 + (\mathbf{\Sigma}_{\text{src}  \rightarrow \text{ref}}^{l,k} - \bm{\mu}_{\mathbf{\Sigma}})^2}{2}
\end{equation}
where $\mathbf{V}$ contains the variance of each covariance element across random data augmentation towards consistent regions on different views. Note that {
% the random data augmentation transformations include the color jittering and Gamma correction.
% The former one generates random variations of brightness, contrast, and saturation. 
% The brightness factor and contrast factor are set to 0.7, and the one of saturation is set to 0.2.
% The latter one randomly samples gamma values within a range of 0.5-2.0 to augment the images.
The random data augmentation transformations include the color jittering and gamma correction.
Color jittering generates random variations of brightness, contrast, and saturation. 
The brightness and contrast factor are set to 0.7, and the one of saturation is set to 0.2.
Gamma correction randomly samples gamma values within a range of 0.5-2.0 to augment the images.
}

As presented in the figure, we then filter out the sensitive element from $\mathbf{V}$ with a threshold $\tau$ using the indicator function: $\mathbb{1} (\mathbf{V} > \tau)$.
$\tau$ is an adaptive threshold calculated following \cite{choi2021robustnet} (Details are discussed in the appendix).
Furthermore, since the covariance matrix is a symmetric matrix, only the upper triangle elements are useful for the whitening loss.
Hence, we can calculate the mask given the upper triangle matrix $\text{triu} (\bm{1})$:
\begin{equation}
	% \small
	\mathbf{O} = \mathbb{1} (\mathbf{V} > \tau) \odot \text{triu}(\bm{1}), \bm{1} \in \mathbb{R}^{C \times C}
\end{equation}
%where $\text{triu} (\bm{1})$ returns an upper triangle matrix.

Finally, the DCW loss is defined as:
\begin{equation}
	% \small
	L_{\text{DCW}}^{l,k} = \mathbb{E} [ \| \Sigma_{\text{ref} \rightarrow \text{src}}^{l,k} \odot \mathbf{O} - \epsilon \|_1 + \| \Sigma_{\text{src} \rightarrow \text{ref}}^{l,k} \odot \mathbf{O} - \epsilon \|_1 ]
\end{equation}
where $0<\epsilon<<1$ is set as a threshold aimming to relax the constraints of whitening loss during optimization.
$\epsilon$ is added to avoid the unexpected failures (NaN values) caused by whitening loss.

\subsubsection{Overall loss}

The final training loss is a weighted sum of depth estimation loss in MVS and the proposed DCW loss:

\begin{equation}
	% \small
	L = L_{\text{depth}} + \frac{\lambda}{(N-1)LK} \sum_{n=1}^{N-1} \sum_{l=1}^{L} \sum_{k=1}^{{K_{\text{clu}}}} L_{\text{DCW}}^{l,k}
	\label{eq18}
\end{equation}
where $L_{\text{depth}}$ is the commonly applied smooth huber loss following \cite{gu2020cascade}.
$\lambda$ is the weight of DCW loss and is set to $1$ in default.
The $N$ views of multi-view images are separated in $N-1$ paired samples comprised of reference view ($n=1$) and source view ($n=2,...,N$).
$L$ layers output standardized features to compute whitening loss, and {$K_{\text{clu}}$} clusters are used as shown in Fig. \ref{fig:framework}.

\section{Experiment}
\label{sec:exp}

\subsection{Datasets}
\label{sec:exp:datasets}

DTU (\textbf{DT}) \cite{jensen2014large} is an indoor MVS dataset.
Each scene is taken from 49 positions and the image resolution is $1600 \times 1200$.
7 different light conditions are included in each scene.
Following the separation of MVSNet \cite{yao2018mvsnet}, 78 scenes are included in the training set, 17 scenes are used in the validation set, and 21 scenes are used in the test set.
% The training set is used as a candidate for source domains, and the test set is used as a candidate for target domains.

BlendedMVS (\textbf{BL}) \cite{yao2020blendedmvs} is a large-scale MVS dataset, which contains various scenes, including architectures, sculptures, and small objects.
Following their official split\footnote{\url{https://github.com/YoYo000/BlendedMVS}}, 106 scenes are used in the training set and 7 scenes are used in the evaluation set.
% The training set is used as a candidate for source domains, and the evaluation set is used as a candidate for target domains.

GTASFM (\textbf{GS}) \cite{wang2020flow} is a synthetic dataset collected from GTA-V, a famous video game with realistic scenarios.
Following the official split, 200 scenes are included in the training set, and 19 scenes are included in the test set.
The original images provided in \textbf{GS} are a sequence of images extracted from a video, paired with camera parameters and ground truth depth maps.
To support the format of MVS networks, these image sequences are firstly processed with COLMAP and the neighboring images are selected based on the similarity score of matching pixels among views.
The format is saved following the format\footnote{\url{https://github.com/YoYo000/MVSNet/blob/master/mvsnet/colmap2mvsnet.py}} defined in the open-source code of MVSNet.
% The training set is used as a candidate for source domains, and the test set is used as a candidate for target domains.

Tanks\&Temples (\textbf{TT}) \cite{knapitsch2017tanks} is a large-scale MVS dataset collected in outdoor scenes with realistic conditions.
It contains 7 scenes in the training set, 8 scenes in the Intermediate evaluation set, and 6 scenes in the Advanced evaluation set.
Following the preprocessed evaluation set provided by MVSNet, we use both the Intermediate and Advanced sets for evaluation.
However, no processed training set is available in the open-source codes of MVS methods.
Hence, we further use COLMAP to sparsely reconstruct scenes on the training set of \textbf{TT}, and separate the neighboring views based on the similarity of matching points in different views.
More importantly, the depth annotations are not provided in the training set of \textbf{TT}, hence we render the densely reconstructed point clouds of all scenes via COLMAP to obtain the depth map on each view.
Our processed training set of \textbf{TT} is used as a candidate for source domains, and the evaluation set provided by MVSNet is used as a candidate for target domain.

PASMVS (\textbf{PA}) \cite{broekman2020pasmvs} is a perfectly accurate synthetic MVS dataset\footnote{\url{https://github.com/andrebroekman/PASMVS}}, consisting of 400 scenes with ground truth depth maps and camera parameters.
Each scene is composed of a unique combination of different camera focal lengths, geometrical models, environmental textures, and materials.
The material properties are primarily specular, which simulate different light conditions with different diffuse materials.
Here we utilize the test set of \textbf{PA} as an extra evaluation dataset while selecting target domains.

\subsection{{MMD Difference among Datasets}}

{
% The visualization result of the MMD distances among different source domains and target domains is shown in Fig. \ref{fig:domain_generalization_task}.
In Fig. \ref{fig:domain_generalization_task}, the MMD distances among different source domains and target domains are visualized to reveal the difference among various MVS datasets.
To calculate the MMD distance, we use a pre-trained VGG network to compute the embedding feature of all images in the domain.
Each image in the MVS training/testing set is converted into a feature vector with 4096 channels.
All these feature vectors are concatenated together to construct a feature matrix describing the corresponding domain.
Then, we can calculate the MMD distance with these feature matrices to assess the distribution gap among different MVS datasets.
}

{
From the presented results in Fig. \ref{fig:domain_generalization_task}, several interesting phenomenons can be found:}
\begin{itemize}
    \setlength{\itemsep}{0pt}
    \setlength{\parsep}{0pt}
    \setlength{\parskip}{0pt}
    \setlength{\topsep}{0pt}
    \setlength{\partopsep}{0pt}
    \item {Compared with the training set of \textbf{DT}, the training set of \textbf{BL} has a much smaller distribution gap towards the evaluation set of \textbf{TT} (\textbf{BL} $\rightarrow$ \textbf{TT}: 62.2; \textbf{DT} $\rightarrow$ \textbf{TT}: 108.1). This explains the fact that the performance of MVS models trained on \textbf{BL} is better than the models trained on \textbf{DT}, which is also revealed by previous works \cite{ding2022transmvsnet, peng2022rethinking}. Furthermore, it means that the impressive generalization results of many recent methods on \textbf{TT} benchmark may not be totally convincing, because the performance improvement may be dependent on the benefit of the large-scale dataset with a small distribution gap like \textbf{BL}, even the joint training of \textbf{BL} and \textbf{DT}. Consequently, we investigate the domain generalization MVS problem in this paper and restrict the setting of generalization with only one single domain to diminish the aforementioned confounding factor.}
    \item {The additional evaluation set of \textbf{PA} has much higher distribution gap compared with other target domains. Note that \textbf{PA} simulates various conditions of object materials and light conditions, which naturally possesses larger difference compared with other target domains. Since \textbf{DT} models 7 different light conditions in each scene, the distribution gap of \textbf{DT} $\rightarrow$ \textbf{PA} is much smaller than the ones of other source domains. Consequently, the illuminance is also an important factor that differs different MVS dataset between \textbf{DT}/\textbf{PA} and \textbf{BL}/\textbf{GS}/\textbf{TT}.}
\end{itemize}

\begin{figure}[t]
  \centering
  \includegraphics[width=0.9\linewidth]{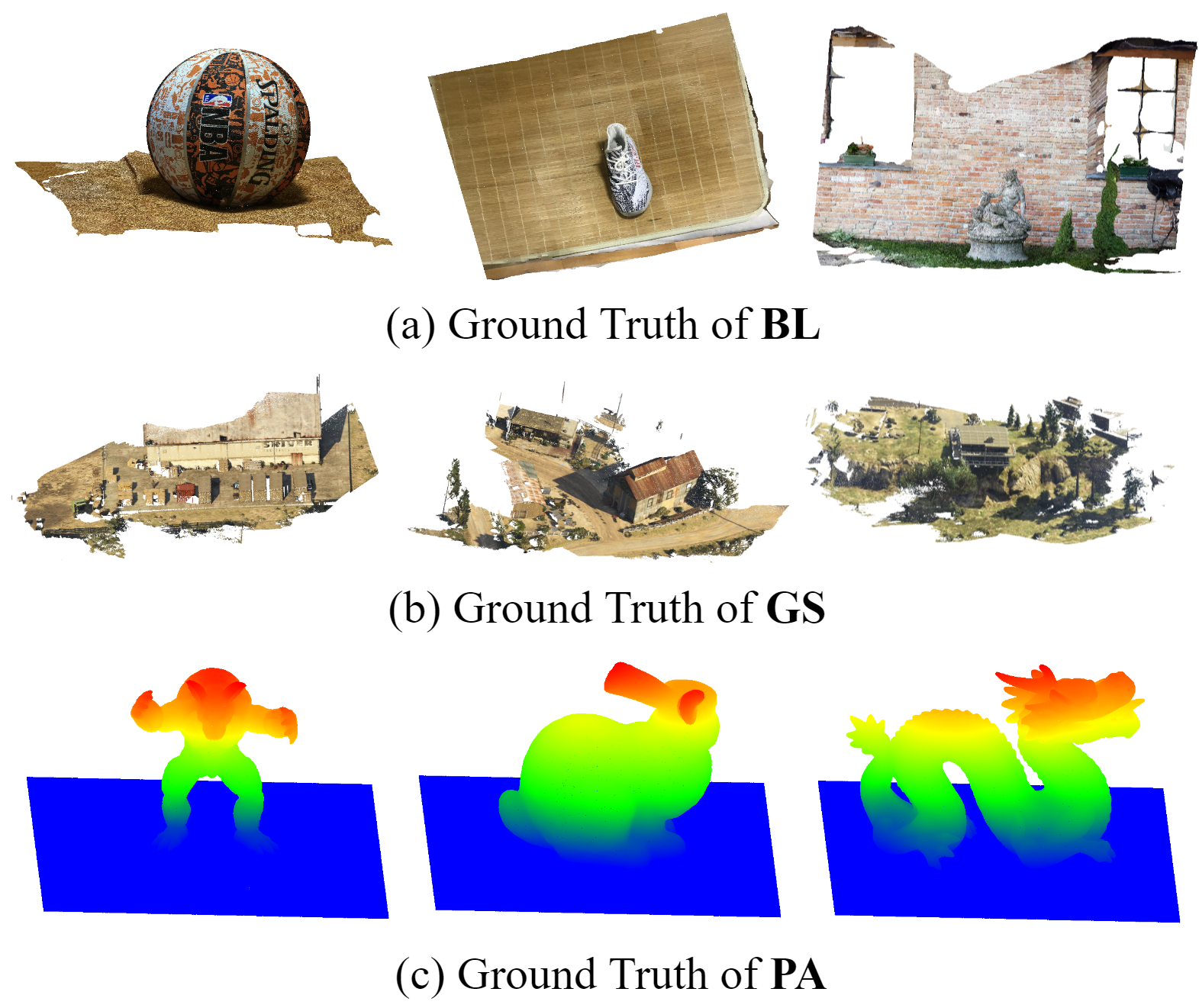}
  \caption{Some examples of the processed ground truths of \textbf{BL}/\textbf{GS}/\textbf{PA}.}
  \label{fig:ground_truth}
\end{figure}

\subsection{3D Ground Truth and Evaluation Benchmark}

{
Although \textbf{DT} and \textbf{TT} contain their own 3D reconstruction evaluation benchmarks, the remaining datasets \textbf{BL}, \textbf{GS}, and \textbf{PA} do not have similar benchmarks for 3D reconstructed results.
Hence, we also need to design customized quantitative benchmarks to measure the performance of 3D reconstruction.
}
To assess the reconstruction quality quantitatively, collecting the 3D ground truth is indispensable for the evaluation of MVS 3D reconstruction.
Hence, we need to prepare the 3D ground truth of each scene in the target domains and propose an equitable evaluation metric to measure the reconstruction results.

\subsubsection{DT Evaluation}

For \textbf{DT}, the ground truth and official evaluation benchmark are provided in the official website\footnote{\url{http://roboimagedata.compute.dtu.dk/?page_id=36}}.
During the evaluation, the metric of \emph{Accuracy}, \emph{Completeness}, and \emph{Overall} are used to measure the reconstruction results.
\emph{Accuracy} is used to measure the average distance from the reconstructed point clouds to the ground truth point clouds.
\emph{Completeness} is used to measure the average distance from the ground truth point clouds to the reconstructed point clouds.
\emph{Overall} is the average of \emph{Accuracy} and \emph{Completeness}, which is used as an overall assessment of the reconstruction quality.

\subsubsection{TT Evaluation}

For \textbf{TT}, the ground truth is not available, and the official evaluation benchmark requires submission of reconstructed point clouds to their official website\footnote{\url{https://www.tanksandtemples.org/}}.
The test set is separated into 2 parts: the \emph{Intermediate} set and the \emph{Advanced} set.
\emph{F-score} on each scene is used to measure the reconstruction quality.

\subsubsection{BL/GS/PA Evaluation}

% xu2022semi
For \textbf{BL}/\textbf{GS}/\textbf{PA}, we have to prepare the ground truth and build the evaluation benchmark by ourselves, since no existing evaluation benchmark is available in the open-source community.

To get the 3D ground truth, we need to process \textbf{BL}/\textbf{GS}/\textbf{PA} differently:
\begin{itemize}
    \setlength{\itemsep}{0pt}
    \setlength{\parsep}{0pt}
    \setlength{\parskip}{0pt}
    \setlength{\topsep}{0pt}
    \setlength{\partopsep}{0pt}
    \item For \textbf{BL}, the provided ground truth mesh is comprised of various splits. 
    The mesh parts are first merged to construct the complete mesh as ground truth.
    Then, we utilize Monte-Carlo sampling to uniformly sample point cloud from the ground truth mesh.
    The sampled point cloud in each scene is used as 3D ground truth in the following evaluation stage.
    Some examples of the sampled ground truth point cloud are shown in Fig. \ref{fig:ground_truth}(a).
    \item For \textbf{GS}, the ideally accurate depth maps and camera parameters are fused with Gipuma \cite{galliani2015massively} to construct the ground truth point cloud.
    It is noted that the extrinsic matrix in GTASFM is the inverse form of the ones defined in Gipuma.
    Some examples of the ground truth are visualized in Fig. \ref{fig:ground_truth}(b).
    \item For \textbf{PA}, the provided ground truth is 3d mesh template without texture.
    We utilize Monte-Carlo sampling to uniformly sample the ground truth point cloud from the original mesh.
    Some examples of the ground truth points are shown in Fig. \ref{fig:ground_truth}(c).
\end{itemize}

To construct the evaluation benchmark, we follow the evaluation protocol defined by \textbf{TT} \cite{knapitsch2017tanks} and \cite{xu2022semi}, which uses \emph{Precision}, \emph{Recall}, and \emph{F-score} to measure the reconstruction quality.

Denote that the ground truth point cloud is $P_G \in \mathbb{R}^{M \times 3}$, and the reconstructed point cloud is $P_R \in \mathbb{R}^{N \times 3}$.
Following the definition of Chamfer Distance (CD), we can have:
\begin{equation}
\begin{aligned}
    D_{CD} = \frac{1}{M} \sum_{p_m \in P_G} \min_{p_n \in P_R} \| p_m - p_n \|_2^2 \\
    + \frac{1}{N} \sum_{p_n \in P_R} \min_{p_m \in P_G} \| p_m - p_n \|_2^2
\end{aligned}
\label{eqs1}
\end{equation}
where $p_m$ is any point of ground truth point cloud $P_G$, and $p_n$ is any point of reconstructed point cloud $P_R$.
We can divide Eq. \ref{eqs1} into 2 terms:
\begin{equation}
    D_{G \rightarrow R} = \frac{1}{M} \sum_{p_m \in P_G} \min_{p_n \in P_R} \| p_m - p_n \|_2^2
    \label{eqs2}
\end{equation}
\begin{equation}
    D_{R \rightarrow G} = \frac{1}{N} \sum_{p_n \in P_R} \min_{p_m \in P_G} \| p_m - p_n \|_2^2
    \label{eqs3}
\end{equation}
where Eq. \ref{eqs2} represents the average distance of all points in the ground truth point cloud $P_G$ towards the reconstructed point cloud $P_R$, which can be viewed as a measure of completeness.
And Eq. \ref{eqs3} represents the average distance of all points in the reconstructed point cloud $P_R$ towards the ground truth point cloud $P_G$, which can be treated as a measure of accuracy.

The calculated distances can further be aggregated to define the recall and precision given a threshold $d$:
\begin{equation}
    \text{Reca}(d) = \frac{100}{M} \sum_{p_m \in P_G} [\min_{p_n \in P_R} \| p_m - p_n \|_2^2 < d]
    \label{eqs4}
\end{equation}
\begin{equation}
    \text{Prec}(d) = \frac{100}{N} \sum_{p_n \in P_R} [\min_{p_m \in P_G} \| p_m - p_n \|_2^2 < d]
    \label{eqs5}
\end{equation}
where $[]$ is the Iverson bracket.
The definition of \emph{Recall} and \emph{Precision} are respectively provided in Eq. \ref{eqs4} and Eq. \ref{eqs5}.

The definition of \emph{F-score} can be calculated from \emph{Recall} and \emph{Precision}:
\begin{equation}
    \text{F-score}(d) = \frac{2 \text{Prec}(d) \text{Reca}(d)}{\text{Prec}(d) + \text{Reca}(d)}
    \label{eqs6}
\end{equation}

Note that \emph{F-score}, \emph{Precision}, and \emph{Recall} are defined to lie in the range of $[0 - 100]$.
The higher these scores are, the better the performance is.

\subsection{Network Details}
\label{sec:exp:impl}

We implement our method in PyTorch and train the models with 4 NVIDIA A-100 GPUs.
The default setting is the same as the backbone network.
For the hyper-parameter settings, $\lambda=1.0$, $K=8$, $\epsilon=0.02$.
%For the hyper-parameter settings, the $\lambda$ in Eq. \ref{eq18} is set to 1.0, and the number of clusters $K$ is set to 8.
The number of adopted layers is set to 3.
We train the network for 24 epochs on \textbf{DT}, \textbf{BL}, \textbf{GS}, and 32 epochs on \textbf{TT}.
Since the training set of \textbf{TT} is smaller than the others, we maintain a fixed range of iterations by increasing the epochs a little.
The number of clusters on depth map is set to 8 in default.
If not mentioned, the adopted setting in training is the same as the one of the backbone MVS network.

\begin{table*}[t]
	\centering
 \caption{Quantitative comparisons with the-state-of-art MVS methods utilizing \textbf{DT} as source domain. Note that all model is purely trained by DTU training set, and tested on other datasets without any finetuning. Best results are in bold, second best are underlined.}
	\resizebox{\hsize}{!}{
		\begin{tabular}{l|cc|c|cc|c|cc|c|cc|c}
			\hline
			\multirow{2}{*}{Methods}   &  \multicolumn{3}{c|}{\textbf{DT $\rightarrow$ DT}}  &  \multicolumn{3}{c|}{\textbf{DT $\rightarrow$ BL}}  & \multicolumn{3}{c|}{\textbf{DT $\rightarrow$ GS}}   & \multicolumn{3}{c}{\textbf{DT $\rightarrow$ PA}}  \\ \cline{2-4} \cline{5-7} \cline{8-10} \cline{11-13}
 			  & {Acc.$^\downarrow$}  & {Comp.$^\downarrow$} & {Overall$^\downarrow$} & Prec.$^\uparrow$  & Reca.$^\uparrow$ & {F-score}$^\uparrow$ & Prec.$^\uparrow$  & Reca.$^\uparrow$  & {F-score}$^\uparrow$ & Prec.$^\uparrow$  & Reca.$^\uparrow$  & {F-score}$^\uparrow$  \\ \hline 
			MVSNet \cite{yao2018mvsnet}             & {0.396} & {0.527} & {0.462} & 29.86 & 25.68 & 22.42  & 16.17 & 18.94 & 16.74  & 20.36 & 24.21 & 21.44    \\
                CasMVSNet \cite{gu2020cascade}           & {0.325} & {0.385} & {0.355} & 48.21 & 39.91 & 39.06  & 42.99 & 36.93 & 38.74  & 42.61 & 41.76 & 40.78    \\
			PatchMatchNet \cite{wang2021patchmatchnet}    & {0.427} & {0.277} & {0.352} & 41.15 & {49.34} & 41.69 & 49.47 & 38.25 & 42.66 & 30.12 & 40.42 & 33.95 \\
			IterMVS \cite{wang2022itermvs}           & {0.373} & {0.354} & {0.363} & 47.21 & 43.93 & 41.79 & 47.36 & 36.91 & 40.92 & 45.00 & 44.84 & 44.37     \\
			MVSTER \cite{wang2022mvster}             & {0.350} & {0.276} & {\textbf{0.313}} & 41.29 & 40.60 & 36.26 & 24.86 & 18.64 & 20.75 & 33.50 & {52.28} & 40.45 \\
			CDS-MVSNet \cite{giang2021curvature}     & {0.352} & {0.280} & {0.316} & 43.74 & 33.97 & 33.89 & 22.45 & 22.71 & 21.66 & 29.09 & 20.92 & 23.72 \\
			UniMVSNet \cite{peng2022rethinking}      & {0.352} & {0.278} & {\underline{0.315}} & 49.09 & 35.06 & 37.93  & {54.30} & 30.66 & 38.47  & {64.46} & 22.07 & 32.24             \\ \hline
			% Ours (CasMVSNet)   & 0.3278 & 0.3652 & 0.3465  & 0.5129 & 0.435 & 0.4227   & 0.456  & 0.4377 & 0.4373  & 0.5564 & 0.4884 & 0.5088              \\
%			RobustMVS     & \textbf{51.29} & 43.50 & \textbf{42.27} & 45.60 & \textbf{43.77} & \textbf{43.73} & 55.64 & 48.84 & \textbf{50.88}              \\
		  \textbf{RobustMVS-Cas}     & {0.328} & {0.365} & {0.346} & {52.07} & 42.91 & \underline{42.72} & 45.45 & {44.08} & \textbf{43.85} & 56.74 & 49.99 & \textbf{52.03} \\
            \textbf{RobustMVS-Iter}        & {0.315} & {0.406} & {0.361} & {49.17} & 43.97 & \textbf{43.13} & 51.32 & 39.11 & \underline{43.83} & 49.93 & 44.87 & \underline{45.89} \\
			\hline
	\end{tabular}}
        % \vspace{+0.1cm}
	% \caption{Quantitative comparisons with the-state-of-art MVS methods utilizing \textbf{DT} as source domain. Note that all model is purely trained by DTU training set, and tested on other datasets without any finetuning.}
	% \vspace{-0.4cm}
	\label{tab:comp_dtu_sota}
\end{table*}

\begin{figure*}[t]
	\centering
        \includegraphics[width=\linewidth]{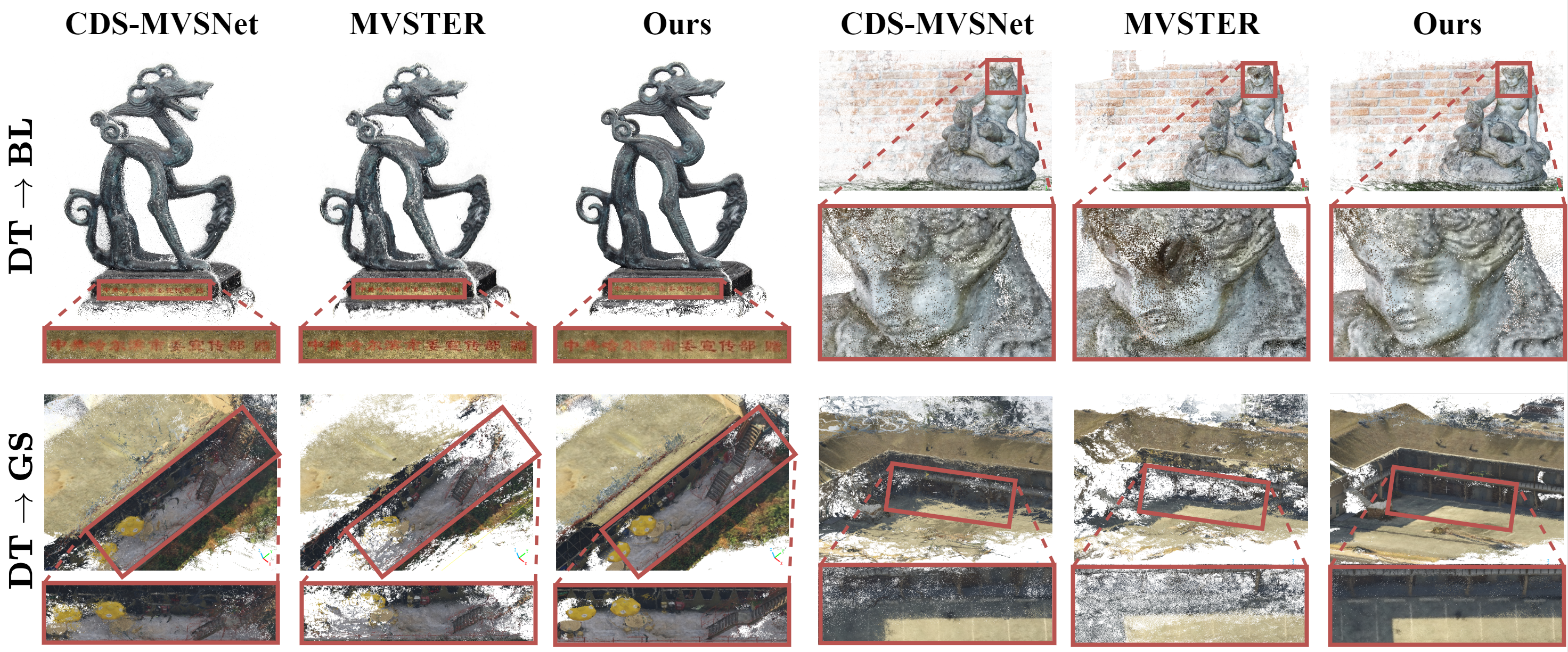}
%	\vspace{-0.6cm}
	\caption{Qualitative comparisons with state-of-the-art MVS methods utilizing \textbf{DT} as source domain.}
	% \vspace{-0.4cm}
	\label{fig:qualitative_dtu_sota_comparison}
\end{figure*}

\begin{table*}[t]
	\centering
 \caption{Quantitative comparison with domain generalization methods utilizing \textbf{DT}/\textbf{BL}/\textbf{GS}/\textbf{TT} as source domain. Best results are in bold.}
	% \resizebox{0.9\hsize}{!}{
        \resizebox{\hsize}{!}{
		\begin{tabular}{l|cc|c|cc|c|cc|c|cc|c}
			\hline
			\multirow{2}{*}{Methods} & \multicolumn{3}{c|}{\textbf{DT $\rightarrow$ DT}}   & \multicolumn{3}{c|}{\textbf{DT $\rightarrow$ BL}}  & \multicolumn{3}{c|}{\textbf{DT $\rightarrow$ GS}}   & \multicolumn{3}{c}{\textbf{DT $\rightarrow$ PA}}    \\ \cline{2-4} \cline{5-7} \cline{8-10} \cline{11-13}
			& Acc.$^\downarrow$  & Comp.$^\downarrow$  & Overall$^\downarrow$ & Prec.$^\uparrow$  & Reca.$^\uparrow$ & F-score$^\uparrow$ & Prec.$^\uparrow$  & Reca.$^\uparrow$  & F-score$^\uparrow$ & Prec.$^\uparrow$  & Reca.$^\uparrow$  & F-score$^\uparrow$   \\ \hline 
			Baseline \cite{gu2020cascade} & 0.325  & 0.385  & 0.355  & 48.21 & 39.91 & 39.06 & 42.99 & 36.93 & 38.74 & 42.61 & 41.76 & 40.78  \\
			FCStereo \cite{zhang2022revisiting}   & 0.3332 & 0.3673 & 0.3503 & 49.80 & 42.35 & 40.76 & 43.09 & 38.99 & 40.42 & 52.88 & 46.73 & 48.44  \\
			IBN \cite{pan2018two}     & 0.3284 & 0.3823 & 0.3554 & 46.22 & 42.35 & 39.52 & \textbf{46.20} & 41.34 & 42.73 & 55.32 & 49.26 & 50.80  \\
			EFDM \cite{zhang2022exact}    & 0.3218 & 0.3746 & 0.3482 & 45.78 & 41.67 & 39.35 & 46.02 & 40.44 & 42.19 & 56.16 & 47.98 & 50.71  \\
			\textbf{RobustMVS-Cas}     & \textbf{0.3209} & \textbf{0.3634} & \textbf{0.3421}  & \textbf{52.07} & \textbf{42.91} & \textbf{42.72} & 45.45 & \textbf{44.08} & \textbf{43.85} & \textbf{56.74} & \textbf{49.99} & \textbf{52.03} \\
			\hline
			
			\multirow{2}{*}{Methods} & \multicolumn{3}{c|}{\textbf{BL $\rightarrow$ DT}} & \multicolumn{3}{c|}{\textbf{BL $\rightarrow$ BL}}   & \multicolumn{3}{c|}{\textbf{BL $\rightarrow$ GS}}   & \multicolumn{3}{c}{\textbf{BL $\rightarrow$ PA}}    \\ \cline{2-4} \cline{5-7} \cline{8-10} \cline{11-13}
			& Acc.$^\downarrow$  & Comp.$^\downarrow$  & Overall$^\downarrow$ & Prec.$^\uparrow$  & Reca.$^\uparrow$ & F-score$^\uparrow$ & Prec.$^\uparrow$  & Reca.$^\uparrow$  & F-score$^\uparrow$ & Prec.$^\uparrow$  & Reca.$^\uparrow$  & F-score$^\uparrow$   \\ \hline 
			Baseline \cite{gu2020cascade} & 0.4343 & 0.6874 & 0.5608 & 44.91 & 37.60 & 36.44 & 52.67 & 55.69 & 53.42 & 35.44 & 30.22 & 31.06   \\
			FCStereo \cite{zhang2022revisiting}    & 0.3534 & 0.3833 & 0.3684 & 44.03 & 43.87 & 38.65 & 53.44 & 55.32 & 53.69 & 37.29 & 31.86 & 32.51   \\
			IBN \cite{pan2018two}      & 0.3490 & 0.3880 & 0.3685 & 44.66 & \textbf{46.73} & 40.54 & 53.67 & 54.95 & 53.66 & 33.80 & 26.36 & 28.03   \\
			EFDM   \cite{zhang2022exact}    & 0.3603 & 0.3844 & 0.3723 & 44.40 & 46.09 & 39.89 & 55.38 & \textbf{56.42} & 55.19 & 34.56 & 30.28 & 30.68   \\
			\textbf{RobustMVS-Cas}      &  \textbf{0.3484} & \textbf{0.3820} &  \textbf{0.3652} & \textbf{47.06} & 46.41 & \textbf{41.82} & \textbf{56.14} & 56.33 & \textbf{55.50} & \textbf{40.03} & \textbf{36.101} & \textbf{36.09}  \\ \hline
			
			\multirow{2}{*}{Methods} & \multicolumn{3}{c|}{\textbf{GS $\rightarrow$ DT}}   & \multicolumn{3}{c|}{\textbf{GS $\rightarrow$ BL}}   & \multicolumn{3}{c|}{\textbf{GS $\rightarrow$ GS}}  & \multicolumn{3}{c}{\textbf{GS $\rightarrow$ PA}}    \\ \cline{2-4} \cline{5-7} \cline{8-10} \cline{11-13}
			& Acc.$^\downarrow$  & Comp.$^\downarrow$  & Overall$^\downarrow$ & Prec.$^\uparrow$  & Reca.$^\uparrow$ & F-score$^\uparrow$  & Prec.$^\uparrow$  & Reca.$^\uparrow$  & F-score$^\uparrow$ & Prec.$^\uparrow$  & Reca.$^\uparrow$  & F-score$^\uparrow$   \\ \hline 
			Baseline \cite{gu2020cascade} & 0.5083 & 0.9305 & 0.7194 & 43.20 & 33.51 & 33.15 & 52.59 & 55.27 & 53.23 & 38.63 & 25.48 & 28.81    \\
			FCStereo \cite{zhang2022revisiting}    & 0.4411 & 0.7490 & 0.5951 & 41.19 & 36.62 & 34.03 & 52.44 & 56.17 & 53.64 & 37.90 & 28.75 & 30.90    \\
			IBN \cite{pan2018two}      & 0.4531 & 0.7195 & 0.5863 & 44.87 & 37.42 & 36.49 & 50.92 & 56.72 & 52.85 & 45.83 & 32.82 & 36.37    \\
			EFDM  \cite{zhang2022exact}     & 0.4084 & 0.5728 & 0.4906 & 44.12 & 38.33 & 36.80 & 51.00 & 56.14 & 52.76 & 34.66 & 27.01 & 28.47    \\
			\textbf{RobustMVS-Cas}      & \textbf{0.3767} & \textbf{0.5390} & \textbf{0.4579} & \textbf{47.78} & \textbf{43.46} & \textbf{41.53} & \textbf{56.38} & \textbf{58.72} &\textbf{ 56.92} & \textbf{48.00} & \textbf{39.59} & \textbf{41.27}    \\ \hline
			
			\multirow{2}{*}{Methods} & \multicolumn{3}{c|}{\textbf{TT $\rightarrow$ DT}}   & \multicolumn{3}{c|}{\textbf{TT $\rightarrow$ BL}}  & \multicolumn{3}{c|}{\textbf{TT $\rightarrow$ GS}}   & \multicolumn{3}{c}{\textbf{TT $\rightarrow$ PA}}    \\ \cline{2-4} \cline{5-7} \cline{8-10} \cline{11-13}
			& Acc.$^\downarrow$  & Comp.$^\downarrow$  & Overall$^\downarrow$ & Prec.$^\uparrow$  & Reca.$^\uparrow$ & F-score$^\uparrow$ & Prec.$^\uparrow$  & Reca.$^\uparrow$  & F-score$^\uparrow$ & Prec.$^\uparrow$  & Reca.$^\uparrow$  & F-score$^\uparrow$   \\ \hline 
			Baseline \cite{gu2020cascade} & 0.4573 & 0.6505 & 0.5539 & 42.80 & 29.70 & 30.83 & 43.74 & 45.41 & 43.67 & 36.59 & 15.62 & 20.67     \\
			FCStereo \cite{zhang2022revisiting}    & 0.3721 & 0.5678 & 0.4700 & 44.52 & 30.02 & 31.17 & 43.53 & 45.80 & 43.83 & 38.27 & 17.64 & 22.79     \\
			IBN \cite{pan2018two}      & 0.4598 & 0.5917 & 0.5257 & 43.91 & 31.91 & 32.99 & 43.92 & 46.01 & 44.23 & 35.57 & 21.51 & 25.46     \\
			EFDM  \cite{zhang2022exact}     & 0.3934 & 0.5635 & 0.4784 & 40.63 & 30.27 & 30.79 & 40.37 & 44.64 & 41.54 & 32.71 & 14.76 & 19.32     \\
			\textbf{RobustMVS-Cas}      & \textbf{0.3687} & \textbf{0.5320} & \textbf{0.4504} & \textbf{47.01} & \textbf{32.45} & \textbf{34.33} & \textbf{46.42} & \textbf{46.95} & \textbf{47.13} & \textbf{43.12} & \textbf{25.20} & \textbf{30.21}     \\
			\hline
	\end{tabular}}
        % \vspace{+0.1cm}
	% \caption{Quantitative comparison with domain generalization methods utilizing \textbf{DT}/\textbf{BL}/\textbf{GS}/\textbf{TT} as source domain.}
	% \vspace{-0.4cm}
	\label{tab:comp_dg}
\end{table*}

\begin{table}[t]
	\centering
 \caption{Generalization results to \textbf{TT} evaluation benchmark.}
	\resizebox{\hsize}{!}{
	\begin{tabular}{l|cc|cc}
		\hline
		\multirow{2}{*}{Methods} & \multicolumn{2}{c|}{\textbf{DT} $\rightarrow$ \textbf{TT}}                          & \multicolumn{2}{c}{\textbf{BL} $\rightarrow$ \textbf{TT}}                          \\ \cline{2-5} 
		& Intermediate$^\uparrow$                      & Advanced$^\uparrow$                       & Intermediate$^\uparrow$                      & Advanced$^\uparrow$                      \\ \hline
		COLMAP \cite{schonberger2016structure}                  & 42.14                     & 27.24                      & 42.14                     & 27.24                     \\ \hline
		MVSNet \cite{yao2018mvsnet}                  & 43.38                     & -           & -          & -          \\
		R-MVSNet \cite{yao2019recurrent}                & 48.4                      & 24.91                      & -          & -          \\
		CIDER \cite{xu2020learning}                   & 46.76                     & 23.12                      & -          & -          \\
		PatchMatchNet \cite{wang2021patchmatchnet}           & 53.15                     & 32.31                      & -          & -          \\
		PatchMatchRL \cite{lee2021patchmatch}            & -          & -           & 51.81                     & 31.78                     \\
		VisMVSNet \cite{zhang2020visibility}               & -          & -           & 60.03                     & 33.78                     \\
		IterMVS \cite{wang2022itermvs}                 & 56.22                     & 33.24                      & 56.94                     & 34.17                     \\
		CasMVSNet \cite{gu2020cascade}               & 56.84 & 31.12 & 57.22 & 32.63 \\
		TransMVSNet \cite{ding2022transmvsnet}             & 55.0                      & 30.2                       & 60.3                      & 35.3                      \\ \hline
		\textbf{RobustMVS-Cas}            & \textbf{57.53}                     & \textbf{32.68}                      & 58.49                     & 33.61                     \\
		\textbf{RobustMVS-Trans}          & 57.43                     & 31.17                      & \textbf{60.80}                     & \textbf{36.66}                     \\ \hline
	\end{tabular}
	}
        % \vspace{+0.05cm}
	% \caption{Generalization results to \textbf{TT} evaluation benchmark.}
	\label{tab:comp_tt}
	% \vspace{-0.45cm}
\end{table}

\subsection{Comparison with MVS methods}

In Table \ref{tab:comp_dtu_sota}, we compare the quantitative performance on domain generalization with other state-of-the-art MVS methods.
Only the training set of \textbf{DT} is selected as source domain to train the model, and we evaluate the performance on other source domains.
We directly utilize the open-sourced pre-trained model of CasMVSNet \cite{gu2020cascade}, PatchMatchNet \cite{wang2021patchmatchnet}, IterMVS \cite{wang2022itermvs}, MVSTER \cite{wang2022mvster}, CDS-MVSNet \cite{giang2021curvature}, and UniMVSNet \cite{peng2022rethinking}, to test on target domains without finetuning.
These pre-trained models are further used for evaulation on unseen datasets on \textbf{DT} $\rightarrow$ \textbf{BL}, \textbf{DT} $\rightarrow$ \textbf{GS}, and \textbf{DT} $\rightarrow$ \textbf{PA}.
{We further provide experimental results on the same dataset (\textbf{DT} $\rightarrow$ \textbf{DT}) to evaluate the performance.}
In the table, RobustMVS-Cas means that the CasMVSNet is selected as the backbone of our RobustMVS framework.
In analogy, RobustMVS-Iter means that the IterMVS is selected as the backbone network.
The metric of F-score reflects the overall performance of 3D reconstruction.
The best results are in bold, and the second best is marked with underline.

An interesting phenomenon can be found that these state-of-the-art models in comparison may fail to generalize to other domains (\textbf{BL}, \textbf{GS}, \textbf{PA}) well even with impressive performance on \emph{DT}.
The reason is that their parameters are highly tuned to fit well-used datasets \emph{DT} or \textbf{TT} with human efforts, resulting in overfitting effect to some extent when a novel unseen domain in involved.
Smaller models like CasMVSNet, PatchMatchNet, and IterMVS suffer less from the overfitting issue.
Comparing with state-of-the-art MVS methods, our RobustMVS-Cas and RobustMVS-Iter achieves better performance on all 3 cross-domain tracks.
As the results show, our proposed method can effectively reduce the performance degradation in unseen novel domains.
Comparing with the baseline method with same backbone (CasMVSNet / IterMVS), our RobustMVS-Cas and RobustMVS-Iter can improve the generalization performance effectively on all 3 cross-domain tracks.
It shows that our proposed RobustMVS can improve the generalization performance when combined with different MVS backbones.
For qualitative results, we provide the visualization comparisons in Fig. \ref{fig:qualitative_dtu_sota_comparison}.
Some examples under \textbf{DT} $\rightarrow$ \textbf{BL} and \textbf{DT} $\rightarrow$ \textbf{GS} are selected for visualization.
The reconstructed point clouds of RobustMVS-Cas is compared with CDS-MVSNet and MVSTER in the figure.
From the figure, we can find that our method can achieve more accurate and complete reconstruction than these state-of-the-art MVS methods.
{Furthermore, the proposed method can achieve slightly better reconstruction performance on RobustMVS-Cas and RobustMVS-Iter compared with the baseline CasMVSNet and IterMVS. 
It shows that our RobustMVS can preserve great performance in the same domain (\textbf{DT} $\rightarrow$ \textbf{DT}) meantime increasing the generalization aility towards other domains (\textbf{DT} $\rightarrow$ \textbf{BL}/\textbf{GS}/\textbf{PA}).}

% From the table, an interesting phenomenon can be found that although the impressive performance on \emph{DT} evaluation benchmark, the state-of-the-art models sometimes may fail to generalize to other domains well, which means they are overfitting to some extent.
% And smaller models like CasMVSNet, PatchMatchNet, and IterMVS suffer less from the overfitting issue.
% From the table, our proposed RobustMVS framework provides superior performance of the overall F-score when generalizing to \textbf{BL}, \textbf{GS}, and \textbf{PA}.
% Furthermore, the qualitative results is presented in Fig. \ref{fig:qualitative_dtu_sota_comparison}.

\subsection{Comparison with DG methods}

\begin{figure*}[t]
	\centering
	\includegraphics[width=0.87\linewidth]{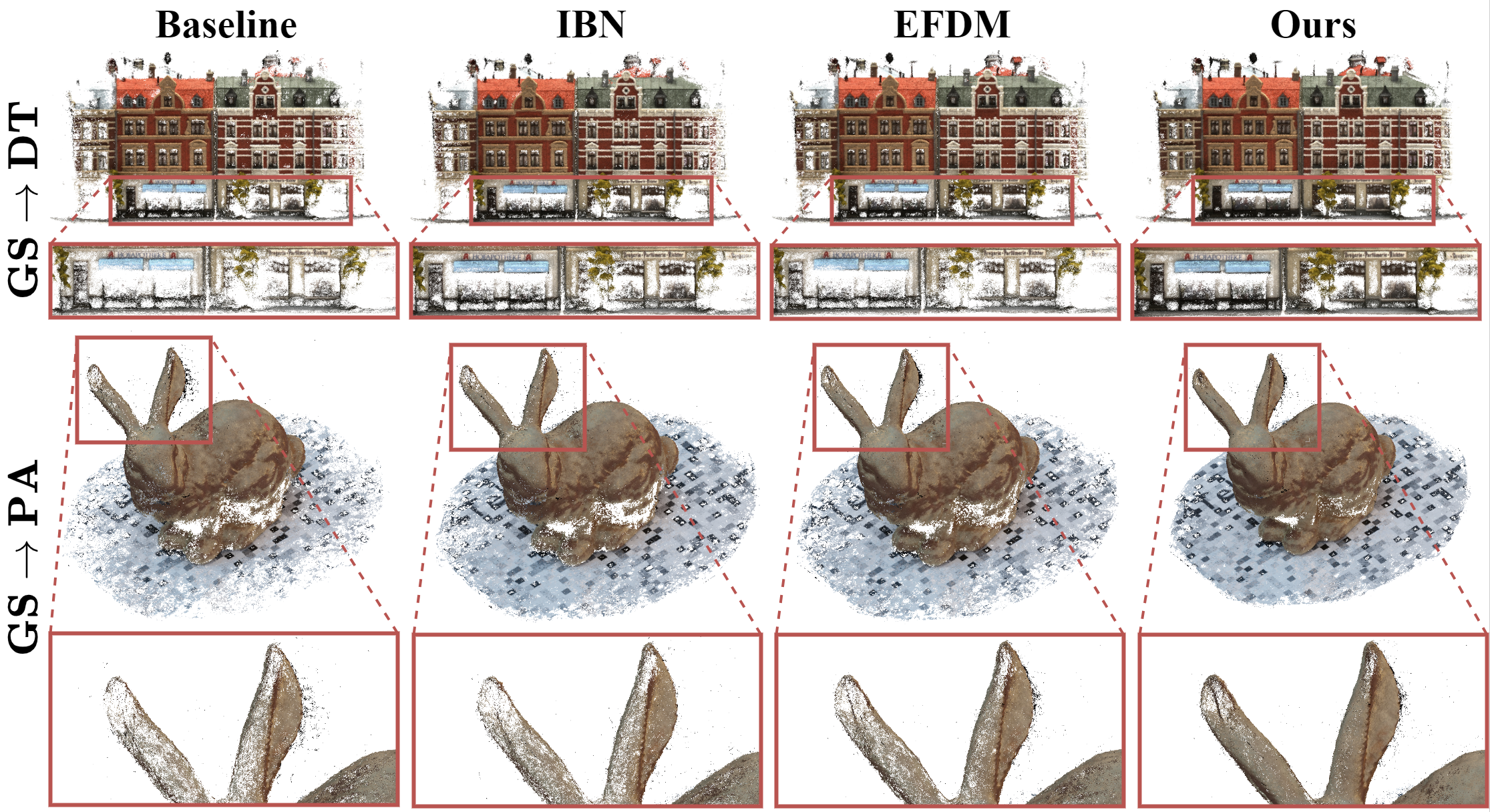}
        % \includegraphics[width=\linewidth]{fig/qualitative_generalization_comparison.png}
	%   \vspace{-0.6cm}
	\caption{Qualitative comparison with domain generalization methods utilizing \textbf{GS} as source domain.}
	% \vspace{-0.4cm}
	\label{fig:qualitative_generalization_comparison}
\end{figure*}

In Table \ref{tab:comp_dg}, we provide the quantitative comparison between our method and other domain generalization methods.
For comparison, we extend the implementation of Stereo-Selective Whitening (SSW) in FCStereo \cite{zhang2022revisiting} to the setting of MVS.
FCStereo is a pioneering work in domain generalized stereo with 2 camera settings, which provides a robust baseline for comparison in the table.
% Following IBN \cite{pan2018two} and EFDM \cite{zhang2022exact}, we modify the backbone of CasMVSNet \cite{gu2020cascade} according to the official instructions.
{% IBN [37] and EFDM [58] are state-of-the-art whitening-based domain generalization techniques that can be plugged into the backbone network. 
IBN [37] and EFDM [58] are state-of-the-art domain generalization modules that can be directly plugged into the backbone network. 
% To provide a direct comparison with these whitening-based methods, we modify the backbone CasMVSNet following the setting of IBN and EFDM to conduct the baselines in Table II.
Since our proposed RobustMVS only insert several instance normalization layers into the backbone network, it can provide a direct and fair comparison between our proposed method and these plug-in domain generalized modules.}
The same backbone network of CasMVSNet \cite{gu2020cascade} is used.
We report the quantitative results of cross-domain generalization MVS on 16 different tracks in the table.
The source domain and target domain is respectively selected from: \textbf{DT}, \textbf{BL}, \textbf{GS}, and \textbf{PA}.
% As the results of F-score on all 16 tracks show, our method outperforms other methods with a distinct margin.
From the direct comparison between FCStereo and our method, our method can significantly improve the reconstruction performance that provides a better multi-view feature-whitening based solution for domain generalization in MVS.
Moreover, our method can also outperform other domain generalized baselines like IBN and EFDM on the metric of \emph{Overall} in \textbf{DT} and \emph{F-score} in \textbf{BL}, \textbf{GS}, and \textbf{PA}.
The qualitative results presents the comparison on some reconstructed samples in Fig. \ref{fig:qualitative_generalization_comparison}.
Some examples of \textbf{GS} $\rightarrow$ \textbf{DT} and \textbf{GS} $\rightarrow$ \textbf{PA} are shown in the figure.
From the figure, it can be found that our method can reconstruct better 3D point cloud compared with other method, and suffer less from the distribution gap when generalizing from synthetic domain \textbf{GS} to realistic domains.

% More implementation details are shown in the appendix.
% Our method achieves better performance compared with other methods on all tracks presented in the table.

% Moreover, qualitative results are shown in Fig. \ref{fig:qualitative_generalization_comparison}.

\begin{table}[t]
% \caption{Ablation results of generalization to \textbf{TT} evaluation benchmark with the same backbone \cite{gu2020cascade}.}
\caption{Improvement of generalization results to \textbf{TT} evaluation benchmark with the same backbone, CasMVSNet \cite{gu2020cascade}.}
\resizebox{\hsize}{!}{
    \begin{tabular}{c|cc|cc}
        \hline
        \multirow{2}{*}{Methods} & \multicolumn{2}{c|}{\textbf{DT} $\rightarrow$ \textbf{TT}} & \multicolumn{2}{c}{\textbf{BL} $\rightarrow$ \textbf{TT}} \\
        & Intermediate$^\uparrow$         & Advanced$^\uparrow$        & Intermediate $^\uparrow$       & Advanced$^\uparrow$        \\ \hline
        Baseline  \cite{gu2020cascade}                 & 56.84                 & 31.12            &  57.22               & 32.63            \\
        RobustMVS-Cas ($\Delta$$^\uparrow$) & +0.69                 & +1.56             & +1.27                & +0.98                 \\ \hline
        \multirow{2}{*}{Methods} & \multicolumn{2}{c|}{\textbf{GS} $\rightarrow$ \textbf{TT}} & \multicolumn{2}{c}{\textbf{TT} $\rightarrow$ \textbf{TT}} \\
        & Intermediate$^\uparrow$         & Advanced$^\uparrow$        & Intermediate $^\uparrow$       & Advanced$^\uparrow$        \\ \hline
        Baseline  \cite{gu2020cascade}                 & 44.65                 & 22.61            &  47.07               & 26.55            \\
        RobustMVS-Cas ($\Delta$$^\uparrow$) & +8.91                 & +7.1             &  +2.85               & +0.95 \\ \hline
    \end{tabular}
}
% \vspace{+0.1cm}
\label{tab:abla_tt}
\end{table}

\begin{table*}[t]
\centering
\caption{Ablation results of different components in RobustMVS.}
\resizebox{\linewidth}{!}{
\begin{tabular}{cccc|cc|c|cc|c|cc|c}
\hline
\multicolumn{4}{c|}{Components} & \multicolumn{3}{c|}{\textbf{DT} $\rightarrow$ \textbf{BL}} & \multicolumn{3}{c|}{\textbf{DT} $\rightarrow$ \textbf{GS}} & \multicolumn{3}{c}{\textbf{DT} $\rightarrow$ \textbf{PA}} \\
Baseline   & DCW   & IN  & DA  & Prec.$^\uparrow$  & Reca.$^\uparrow$  & F-score$^\uparrow$ & Prec.$^\uparrow$  & Reca.$^\uparrow$  & F-score$^\uparrow$ & Prec.$^\uparrow$  & Reca.$^\uparrow$  & F-score$^\uparrow$ \\ \hline
 $\checkmark$          &                   &                 &                 & 48.21            & 39.91            & 39.06                          & 42.99             & 36.93            & 38.74             & 42.61            & 41.76   & 40.78    \\
 $\checkmark$          & $\checkmark$      &                 &                 & 48.20            & 41.44            & 39.70                          & 44.33             & 39.28            & 40.65             & 51.31            & 46.49   & 47.51    \\
 $\checkmark$          & $\checkmark$      & $\checkmark$    &                 & 51.80            & 42.07            & 41.94                          & \textbf{47.10}    & 37.69            & 41.03             & 47.04            & \textbf{57.95}   & 51.55    \\
 $\checkmark$          & $\checkmark$      & $\checkmark$    & $\checkmark$    & \textbf{52.07}   & \textbf{42.91}   & \textbf{42.72}    & 45.45             & \textbf{44.08}   & \textbf{43.85}    & \textbf{56.74}   & 49.99   & \textbf{52.03}    \\ \hline
\end{tabular}}
% \caption{Ablation results of different components in RobustMVS, including the Depth-clustering whitening (DCW) loss, Instance Normalization (IN), Data-augmentation (DA).}
% \caption{Ablation results of different components in RobustMVS.}
% \vspace{-0.3cm}
\label{tab:ablation_components}
% \addtocounter{Stable}{1}
\end{table*}

\begin{table*}[t]
\caption{Ablation experiments of the number of depth clusters $K$. \textbf{GS} (synthetic data) is used as source domain.}
	\resizebox{\linewidth}{!}{
		\begin{tabular}{c|ccc|ccc|ccc|ccc}
			\hline
			\multirow{2}{*}{Depth clusters} & \multicolumn{3}{c|}{\textbf{GS $\rightarrow$ DT}} & \multicolumn{3}{c|}{\textbf{GS $\rightarrow$ BL}}   & \multicolumn{3}{c|}{\textbf{GS $\rightarrow$ GS}}   & \multicolumn{3}{c}{\textbf{GS $\rightarrow$ PA}}    \\ \cline{2-4} \cline{5-7} \cline{8-10} \cline{11-13}
			& Acc.$^\downarrow$  & Comp.$^\downarrow$  & Overall$^\downarrow$ & Prec.$^\uparrow$  & Reca.$^\uparrow$ & F-score$^\uparrow$ & Prec.$^\uparrow$  & Reca.$^\uparrow$  & F-score$^\uparrow$ & Prec.$^\uparrow$  & Reca.$^\uparrow$  & F-score$^\uparrow$   \\ \hline \hline
			K=0                          & 0.4411 & 0.7490 & 0.5951 & 41.19 & 36.62 & 34.03 & 52.44 & 56.17 & 53.64 & 37.90 & 28.75 & 30.90   \\ \hline
			K=4                          & 0.4148 & 0.4148 & 0.5122 & 44.92 & 40.00 & 38.53 & 54.02 & 57.91 & 55.27 & 39.72 & 33.59 & 34.63   \\
			K=6                          & 0.3886 & 0.6315 & 0.5101 & 46.51 & 42.28 & 40.40 & 55.51 & 58.53 & 56.41 & 42.27 & 36.11 & 37.44   \\
			K=8                          & \textbf{0.3767} & \textbf{0.5390} & \textbf{0.4579} & \textbf{47.78} & \textbf{43.46} & \textbf{41.53} & \textbf{56.38} & \textbf{58.72} & \textbf{56.92} & \textbf{48.00} & \textbf{39.59} & \textbf{41.27}   \\ 
			K=10                         & 0.4087 &	0.5993 & 0.5040 & 43.51 & 38.24 & 36.19 & 52.64 & 56.07 & 53.67 & 42.11 & 33.25 & 35.34
			\\
			K=12                         & 0.4141 &	0.6085 & 0.5113 & 41.00 & 38.24 & 35.62 & 52.45	& 57.08	& 54.03	& 33.45	& 30.64	& 30.88
			\\
			K=14                         & 0.4372 & 0.5953 & 0.5162 & 40.93 & 36.64 & 34.35 & 50.78 & 56.34 & 52.71 & 33.84 & 26.88 & 27.89
			\\
			%			K=16                         & 0.4282 &	0.5558 & 0.4920 & 43.55 & 39.78	& 37.75	& 52.33	& 57.31	& 54.03	& 39.64	& 34.40 & 	35.08 \\
			\hline
	\end{tabular}}
	% \caption{Ablation experiments of the number of depth clusters $K$. \textbf{GS} (synthetic data) is used as source domain.}
	% \vspace{-0.4cm}
	\label{tab:abla_depth_cluster}
\end{table*}

\subsection{Generalization on Tanks\&Temples}

The generalization results on \textbf{TT} evaluation set are provided in Table \ref{tab:comp_tt}.
F-score is used to measure the performance of reconstruction results.
The intermediate and advanced sets of \textbf{TT} are both utilized for evaluation following previous works \cite{wang2022itermvs, ding2022transmvsnet}.
The compared methods include: COLMAP \cite{schonberger2016structure}, MVSNet \cite{yao2018mvsnet}, R-MVSNet \cite{yao2019recurrent}, CIDER \cite{xu2020learning}, PatchMatchNet \cite{wang2021patchmatchnet}, PatchMatchRL \cite{lee2021patchmatch}, VisMVSNet \cite{zhang2020visibility}, IterMVS \cite{wang2022itermvs}, CasMVSNet \cite{gu2020cascade}, and TransMVSNet \cite{ding2022transmvsnet}.
The open-sourced pretrained model of the compared methods are utilized for evaluation. 
RobustMVS-Cas and RobustMVS-Trans respectively adopts CasMVSNet and TransMVSNet as the backbone network in our RobustMVS framework.
% {Since TransMVSNet in Table \ref{tab:comp_tt} is trained on multiple datasets (\textbf{DT} + \textbf{BL}) in their official implementations, we use their official code to train the network on single domains respectively and evaluate the generalization performance.}
% {In Table \ref{tab:comp_tt}, }
% For example, the TransMVSNet in Table \ref{tab:comp_tt} 
% Since some compared methods adopts multiple datasets (\textbf{DT} + \textbf{BL}) in their official implementations, we use the official implementation to train the network with only single domain and evaluate the generalization performance.
% In the table, we report the performance of \textbf{DT} $\rightarrow$ \textbf{TT} and \textbf{BL} $\rightarrow$ \textbf{TT} under the setting of single source domain for a fair comparison.
{The criteria for choosing thse networks as baselines is to compare the single domain generalization performance in a comparative setting. Since TransMVSNet adopts multiple datasets (\textbf{DT} + \textbf{BL}) for training, we use their official implementation to retrain the network on single domain respectively to ensure a fair comparison in the single domain generalization benchmark.
In the table, the performance of \textbf{DT} $\rightarrow$ \textbf{TT} and \textbf{BL} $\rightarrow$ \textbf{TT} are reported for comparison.
}
As the results show, our RobustMVS-Trans outperforms the baseline TransMVSNet on the intermediate and advanced set in both 2 settings.
Superior results can also be witnessed in the comparison between RobustMVS-Cas and the baseline CasMVSNet.
When compared with other methods, our RobustMVS can still have competitive performance and even better results.

In Table \ref{tab:abla_tt}, we further provide the improvement of generation results towards \textbf{TT} dataset in evaluation.
The same backbone of CasMVSNet is adopted here.
The evaluation includes 4 different tracks: \textbf{DT} $\rightarrow$ \textbf{TT}, \textbf{BL} $\rightarrow$ \textbf{TT}, \textbf{GS} $\rightarrow$ \textbf{TT}, and \textbf{TT} $\rightarrow$ \textbf{TT}.
It can be explored that the RobustMVS can improve the performance on all these benchmarks in both the intermediate and advanced set.
Furthermore, when the synthetic dataset \textbf{GS} is used as source domain, significant improvement can be witnessed on intermediate set (+8.91) and advanced set (+7.1).

% The performance improvement on each source domain of \textbf{DT}, \textbf{BL}, \textbf{GS}, and \textbf{TT} demonstrates the superior performance of the proposed method.

\begin{figure*}[t]
	\centering
	\includegraphics[width=\linewidth]{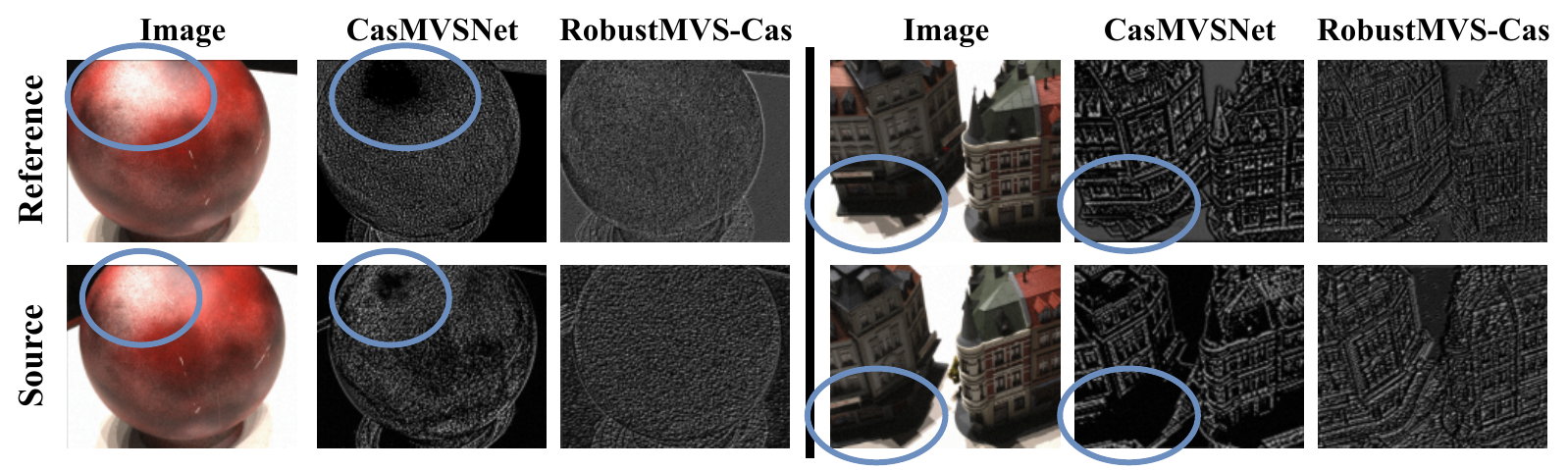}
	%   \vspace{-0.6cm}
	\caption{Grad-cam \cite{selvaraju2017grad} visualization example of RobustMVS under domain generalization setting of \textbf{GS} $\rightarrow$ \textbf{DT}.}
	% \vspace{-0.4cm}
	\label{fig:grad_cam_visualization}
\end{figure*}

\subsection{Ablation Study}

1) \emph{Ablation Analysis of Different Components}: To evaluate the effectiveness of each component in the proposed RobustMVS framework, we conduct ablation experiments on them and provide the quantitative results in Table \ref{tab:ablation_components}.
In the table we present the results on 3 different tracks: \textbf{DT} $\rightarrow$ \textbf{Bl}, \textbf{DT} $\rightarrow$ \textbf{GS}, and\textbf{DT} $\rightarrow$ \textbf{PA}.
The backbone network is CasMVSNet in default.
The "baseline" represents that the original CasMVSNet network.
"DCW" means the proposed DCW-loss is appended in the training phase.
"IN" and "DA" respectively represents the instance normalization module in the network and data augmentation strategy during training.
As the table shows, when DCW-loss is appended to the MVS network training, the F-score is improved on all 3 benchmarks.
Since the Instance Normalization is usually paired with the whitenning loss, better performance can be achieved when IN is also used.
The involvement of data-augmentation strategy enables the feature whitening loss to reduce the domain bias under more various scenarios compared with the naive setting.
As the table shows, each component is effective in improving the generalization performance.

2) \emph{Ablation Analysis of Different Clusters}: To explore the effectiveness of the depth clusters $K$ in DCW loss, we also conduct ablation experiments for different settings: ${K_{\text{clu}}}=0/4/6/8/10/12/14$ on the benchmark of  \textbf{GS} $\rightarrow$ \textbf{DT}, \textbf{GS} $\rightarrow$ \textbf{BL}, \textbf{GS} $\rightarrow$ \textbf{GS}, and \textbf{GS} $\rightarrow$ \textbf{PA}.
When ${K_{\text{clu}}}=0$, it represents the basic feature whitening loss used in FCStereo \cite{zhang2022revisiting}.
When ${K_{\text{clu}}}$ is larger than 0, it represents the depth cluster used in our DCW-loss.
As shown in Table \ref{tab:abla_depth_cluster}, the performance improves when ${K_{\text{clu}}}$ is smaller than 8 but degrades when ${K_{\text{clu}}}$ becomes larger than 8.
The reason is that ${K_{\text{clu}}}$ is small, increasing the number of clusters can regularize the cross-view correspondence in a more detailed local region.
However, when ${K_{\text{clu}}}$ becomes too large, the pixels located in the local corresponding regions contains less variance of style information, which tends to have smaller difference among each other.
Thus the regularization on these small regions can not effectively excavate the style information when calculating feature whitening loss, and degrade the reconstruction performance.

{
There are a few suggestions if you want to adjust the depth cluster number ($K_{\text{clu}}$) when the image resolution (H, W) changes:
a) If the image resolution is similar in scale to the default setting in our synthetic domain ($640 \times 480$ in \textbf{GS}), you can use the default setting of ($K_{\text{clu}}$), i.e. \textbf{DT} ($640 \times 512$).
b) If the difference between the image resolutions of the target dataset and the default synthetic dataset \textbf{GS} is too large, we suggest downsampling the image resolution to a comparative scale of the default one. For example, the image resolution of \textbf{TT} can be resized to $640 \times 480$ from its original one for training.
c) The depth cluster number choice may also be affected by the variety of depth values in a scene (synthetic or real). This is because the balanced distance-based metric in K-Means may be affected by the sparsity of point clouds. For example, for some sharp shapes vertical to the image plane, the sampled points from depth might be pretty sparse. In point clouds, K-Means clustering might ignore these sparse points and focus on the dense points with flat shapes horizontal to the image plane. This problem might occur when the image resolution is relatively small. We can mitigate it by not shrinking the image resolution.
d) We suggest not setting $K_{\text{clu}}$ to be larger than 8 as training might become unstable and return NaN values caused by the strong whitening constraints. Finally, due to the implementation issue in parallelism, we utilize an even number of depth clusters for parallelizing the computation.
}

3) \emph{Grad-cam visualization of MVS}: To explain the effectiveness of the proposed RobustMVS framework, we provide the Grad-cam \cite{selvaraju2017grad} visualization results in Fig. \ref{fig:grad_cam_visualization}.
We compare the baseline model of CasMVSNet and RobustMVS-Cas for visualization.
In default the source domain is synthetic dataset \textbf{GS} and the target domain is \textbf{DT}.
The error between the estimated depth map of the MVS network and the ground truth depth is back-propagated to the convolutional layers, we utilize Grad-cam \cite{selvaraju2017grad} to visualize the gradient on feature maps.
In the first column, the original images of the reference view and source view are shown.
The second and third column respectively show the activation map of baseline model and our proposed method.
The blue circle in the figure points out the ignored activation areas in the baseline model.
For example, in the left example, the light exposure results in textureless regions on the red ball, which has never been seen before in the synthetic dataset \textbf{GS}.
Consequently, the baseline model can not handle this case and these areas are not activated in blue circles.
However, our RobustMVS can provide more complete activation map compared with baseline model on the whole activation map.
From these examples, we can find that our RobustMVS provide a robust regularization to reduce the domain bias and maintain a well-distributed feature activation map across domains.

% \begin{table}[t]
% \centering
% \caption{{Comparison of computation efficiency. The inference time is evaluated on a single A100 GPU.}}
% \resizebox{0.8\hsize}{!}{
% 	\begin{tabular}{c|c|c|c}
% 		\hline
% 		Methods         & Params & GFLOPS   & Inference Time (s) \\ \hline \hline
% 		CasMVSNet & 934.304K     & 268.540G & 0.1904              \\ \hline
% 		RobustMVS-Cas   & 934.192K     & 268.401G & 0.1977              \\ \hline
% 	\end{tabular}
% }
% \label{tab:computation_efficiency}
% \end{table}

\begin{table}[t]
\centering
\caption{{Comparison of computation efficiency. The inference time is evaluated on a single A100 GPU.}}
\resizebox{0.8\hsize}{!}{
	\begin{tabular}{c|c|c|c}
		\hline
		Methods         & Params & GFLOPS   & Inference Time (s) \\ \hline \hline
		CasMVSNet & 934.304K     & 268.540G & 0.1904              \\ \hline
		RobustMVS-Cas   & 934.192K     & 268.401G & 0.1977              \\ \hline
	\end{tabular}
}
\label{tab:computation_efficiency}
\end{table}

{
4) \emph{Computation efficiency}: 
We have compared the baseline of CasMVSNet with our RobustMVS-Cas, and the results are presented in Table \ref{tab:computation_efficiency}. Both networks use the same backbone, but our RobustMVS-Cas replaces some BN layers with IN layers. Our RobustMVS-Cas has slightly fewer parameters and lower GFLOPS than the baseline backbone. However, the inference time of RobustMVS-Cas might be slightly slower than the baseline backbone network. This is because the original backbone uses batch normalization layers that can store the running mean and variance values, reducing computation during inference. In contrast, our RobustMVS-Cas requires extra online computation during inference due to the replacement of some batch normalization layers with instance normalization layers. Despite the slightly slower inference time than the baseline, our RobustMVS-Cas can significantly improve the generalization performance.
% As shown in Table \ref{tab:computation_efficiency}, we present the comparison between the baseline of CasMVSNet and our RobustMVS-Cas. 
% Note that the same backbone are utilized, and our RobustMVS-Cas only changes several BN layers of the original network to IN layers.
% As the table shows, our RobusMVS has slightly smaller number of parameters and GFLOPS than the baseline backbone.
% However, the inference time of RobustMVS might be slighter slower than the baseline backbone network.
% The reason is that the original backbone utilizes the batch normalization layers which can store the running mean and variance values, and reduce the computation in the inference.
% Whereas our RobustMVS replaces some batch normalization layers with instance normalization layers, it requires extra online computation in the inference.
% As a result, despite the slightly slower inference time than baseline, our RobustMVS can apparently boost the generalization performance.
}

% \section{}

% The reason is that when $K$ becomes too large, the pixels in the local regions is too sparse and locate on the boundary, thus degrading the representation of the local regions.

% In Table \ref{tab:abla_tt}, we report the generalization results to \textbf{TT} of the same backbone network \cite{gu2020cascade} trained with and without our RobustMVS framework.
% On all generalization benchmarks, our proposed method can achieve improvement compared with the baseline, especially on the benchmark of synthetic to real ($\textbf{GS} \rightarrow \textbf{TT}$).

%We perform ablation experiments on the effectiveness of the depth clusters $K$ in DCW loss.
%The experiment results are shown in Table \ref{table4}.
%As shown in the table, after applying the DSW loss, the overall performance on \textbf{GS} $\rightarrow$ \textbf{DT}, and the F-score on \textbf{GS} $\rightarrow$ \textbf{BL}, \textbf{GS} $\rightarrow$ \textbf{GS}, \textbf{GS} $\rightarrow$ \textbf{PA}, are improved significantly compared with the baseline ($K=0$).
%By increasing the number of clusters, the performance is increased accordingly as well.

%\subsection{Computational Cost}
%
%In Table \ref{table5}, we report the number of parameters, computation cost, and inference time.
%The compared methods share the same network backbone.
%Compared with the original baseline, our proposed method performs whitening transformation with little computation burden but meantime achieving impressive domain generalization performance.

\section{Conclusion}

% This paper first introduces the domain generalization problem in the field of MVS, and proposes a novel domain-generalized framework, RobustMVS.
% It requires little modifications to the backbone MVS network and applies a Depth-Clustering-guided Whitening (DCW) loss to diminish the domain-specific styles on spatially neighboring regions.
% The experimental results reveal the robust performance of the proposed method in domain generalization benchmark of MVS.

This paper first introduces the domain generalization problem in the field of MVS.
As the visualization of the activation map on MVS networks across different domains shows, the bottleneck lies in the robustness towards cross-view matching features across domain-specific conditions.
Consequently, we propose a novel domain generalized MVS framework, RobustMVS.
It requires little modifications to the backbone MVS network and applies a Depth-Clustering-guided Whitening (DCW) loss to diminish the domain-specific styles on spatially neighboring regions.
% We build domain generalized evaluation benchmarks on 5 datasets and compare the proposed method with existing MVS methods and domain generalization methods.
{We extend the MVS evaluation benchmarks to 5 datasets for domain generalization and compare the proposed method with existing MVS methods and domain generalization methods.}
The experimental results reveal the robust performance of the proposed method in the domain generalization benchmark of MVS.
The final visualization of activation maps in our proposed method gives clues to indicate whether it can reduce the influence of domain bias.
For future work, we will explore the further extension of Transformer-based MVS networks in the domain generalized MVS tasks and try to conduct domain generalization from large-scale synthetic MVS datasets towards realistic MVS datasets.

\bibliographystyle{IEEEtran}
\bibliography{reference.bib}

\vspace{-0.8cm}

\begin{IEEEbiography}
[{\includegraphics[width=1in,height=1.25in,clip,keepaspectratio]{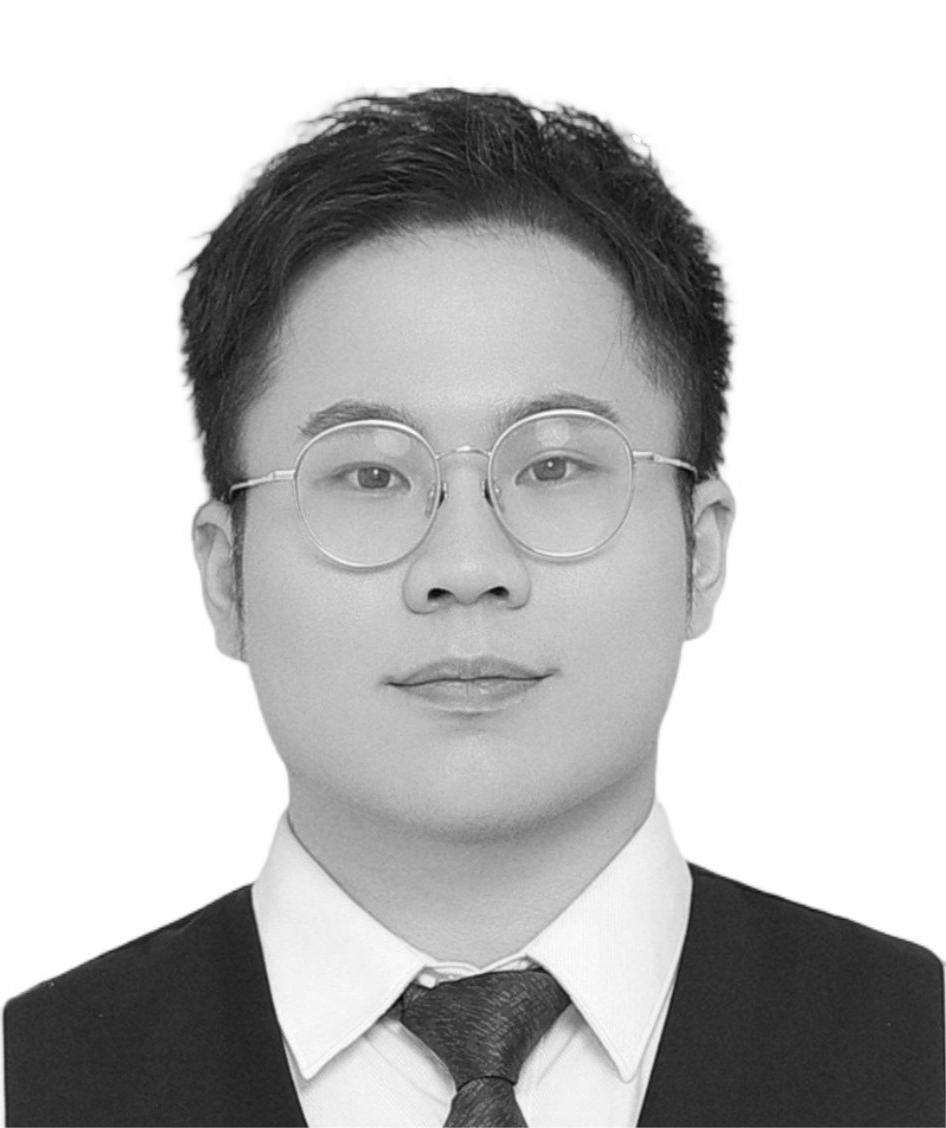}}]{Hongbin Xu}
received B.E. degree from Nanchang University in 2018 and M.S. degree from South China University of Technology in 2021.
He is currently pursuing the Ph.D. degree with the School of Automation Science and Engineering, South China University of Technology, Guangzhou, China. 
His research interests include 3D vision, 3D reconstruction, and computer vision.
\end{IEEEbiography}

\vspace{-0.8cm}

\begin{IEEEbiography}[{\includegraphics[width=1in,height=1.25in,clip,keepaspectratio]{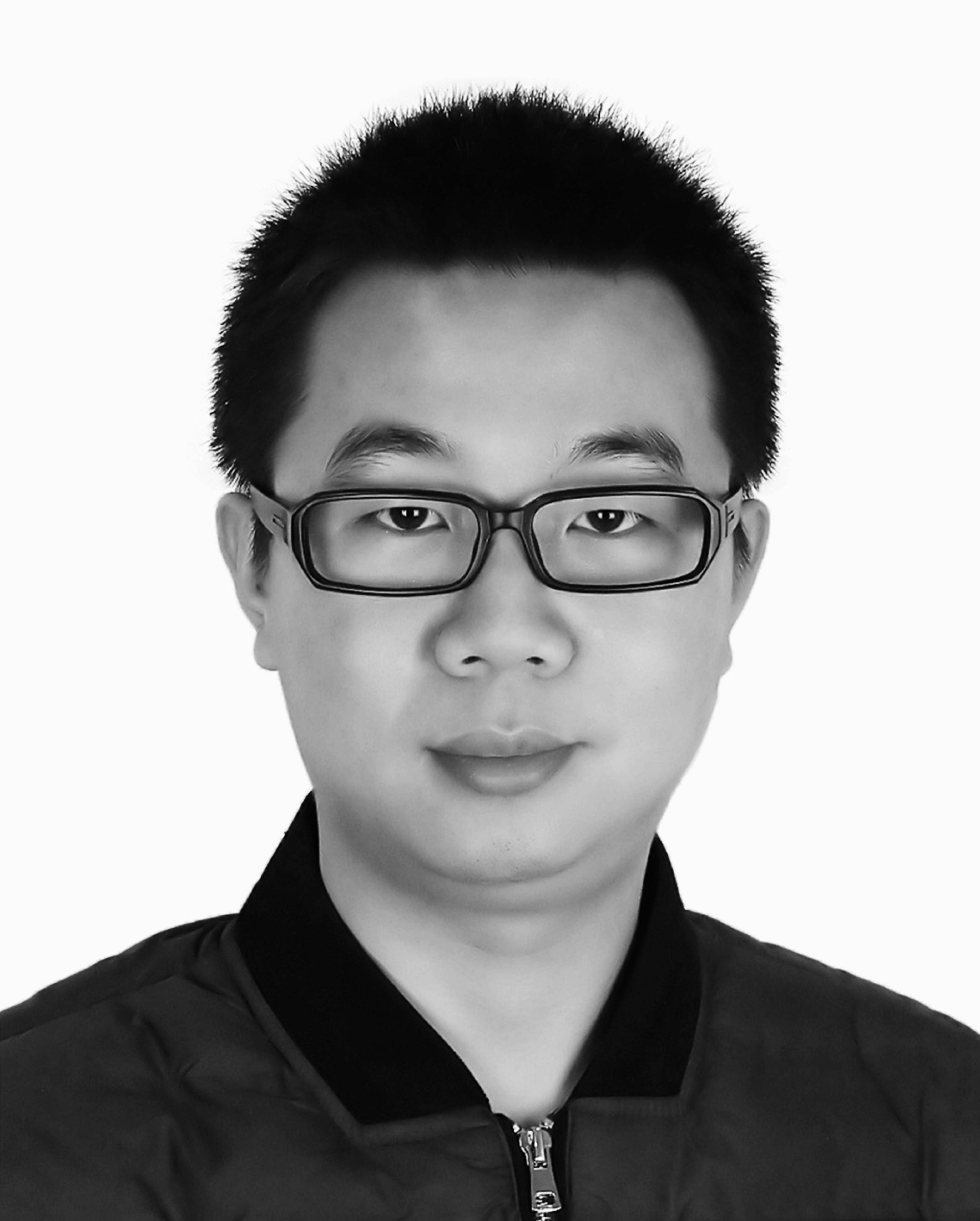}}]{Weitao Chen}
Weitao Chen received the M.S degress from Xiamen University, Xiamen, China, in 2016. He is working as a staff/senior algorithm engineer at Tongyi Laboratory, Alibaba Group. His research interests include 3D vision, AIGC, AI4Science and remote sensing.
\end{IEEEbiography}

\vspace{-0.8cm}

\begin{IEEEbiography}[{\includegraphics[width=1in,height=1.25in,clip,keepaspectratio]{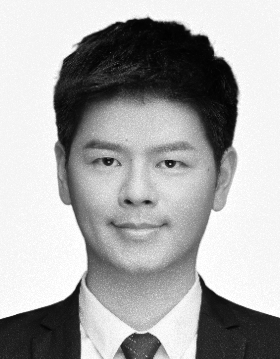}}]{Baigui Sun}
Baigui Sun received the M.Eng. degree from Zhejiang University, Hangzhou, China, in 2014. He joined Alibaba Group, Hangzhou, China, as a computer vision algorithm researcher, in 2014. His research interests include face detection/recognition, deep metric learning, domain adaption, zero/few-shot learning, semi-supervised learning, MVS and computer vision applications. He has published over 26+ articles in international journals and conferences.
\end{IEEEbiography}

\vspace{-0.8cm}

\begin{IEEEbiography}[{\includegraphics[width=1in,height=1.25in,clip,keepaspectratio]{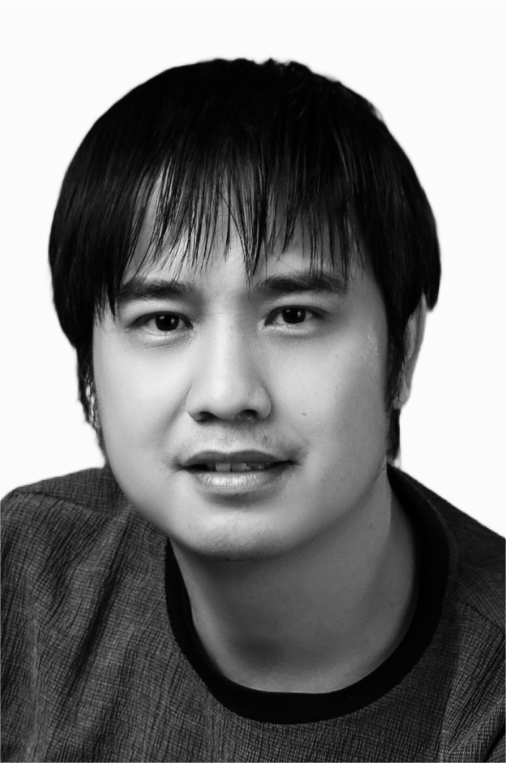}}]{Xuansong Xie}
Xuansong Xie received the PH.D. degree from Jilin University, Jilin, China. He is a senior algorithm expert at Alibaba, serving as the technical director of the Open Vision Intelligence Department at the Tongyi Laboratory. He is also a core member of the Alibaba Group's visual technology team and responsible for the Alibaba Cloud Vision Intelligence Open Platform. His research interests include visual generation, editing, perception and understanding.
\end{IEEEbiography}

\vspace{-0.8cm}

\begin{IEEEbiography}[{\includegraphics[width=1in,height=1.25in,clip,keepaspectratio]{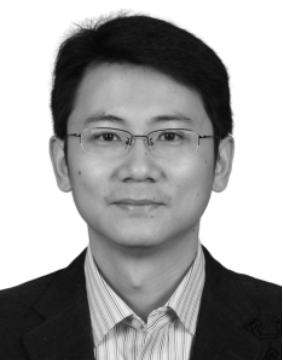}}]{Wenxiong Kang}
(Member, IEEE) received the
M.S. degree from Northwestern Polytechnical University, Xi’an, China, in 2003, and the Ph.D. degree
from the South China University of Technology,
Guangzhou, China, in 2009. He is currently a Professor with the School of Automation Science and
Engineering, South China University of Technology.
His research interests include biometrics identification, image processing, pattern recognition, and
computer vision.
\end{IEEEbiography}

\end{document}